\documentclass[11pt]{article}
\usepackage[T1]{fontenc}
\usepackage[utf8]{inputenc}
\DeclareUnicodeCharacter{0166}{\mbox{\rlap{\raisebox{.35ex}{\hspace{.08em}\rule{.45em}{.035em}}}T}}
\usepackage{lmodern}
\usepackage{amsmath,amssymb,amsthm,graphicx,booktabs,geometry,natbib}
\usepackage{float}
\usepackage{color}
\newcommand{\reviewchange}[1]{#1}
\newcommand{\roundchange}[1]{#1}
\newcommand{\currentchange}[1]{#1}
\newcommand{\thisrevision}[1]{#1}
\newcommand{\layoutrevision}[1]{#1}
\usepackage{hyperref}
\DeclareRobustCommand{\dgpref}[1]{\citet{#1}}
\hypersetup{colorlinks=true,linkcolor=red,citecolor=blue,urlcolor=blue}
\newcommand{\E}{\mathbb{E}}
\newcommand{\ind}{\mathbf{1}}
\providecommand{\keywords}[1]{{\footnotesize\textbf{\textit{Keywords---}} #1}}
\newtheorem{assumption}{Assumption}[section]
\newtheorem{proposition}{Proposition}
\newtheorem{theorem}[proposition]{Theorem}
\newtheorem{corollary}{Corollary}
\theoremstyle{remark}
\newtheorem{remark}{Remark}
\title{\bf MediPFN: Amortized In-Context Nuisance Learning and Inference for Causal Mediation Analysis}
\author{Shi Bo and Jiaming Hu\\\small Department of Mathematics and Statistics, Boston University}
\date{}
\begin{document}
\maketitle
\begin{abstract}
 Causal mediation analysis decomposes a treatment effect into indirect pathways through mediators and direct pathways through other processes.  Tabular foundation models perform classification and regression through in-context learning, providing estimates of   conditional means and propensity scores used  in causal mediation analysis.   Constructing confidence intervals from these estimates requires accounting for uncertainty introduced by fitting and combining several nuisance functions. We introduce MediPFN, an amortized inference framework that adapts a pretrained tabular foundation model to nuisance prediction and combines its predictions through a cross-fitted influence-function estimator. A separate head learns quantiles of the estimator's standardized error from simulated tables with known indirect effects. Supervised fine-tuning, symmetrization, and independent calibration then produce a confidence interval.   We  establish coverage bounds for the resulting intervals   under standard assumptions. Across experiments, MediPFN attains coverage closer to the nominal level than several   comparators. It also reduces the computational cost of uncertainty quantification, running 59.1 times faster than   bootstrap  in matched comparisons.
\end{abstract}

\newpage
\begingroup
\hypersetup{linkcolor=black}
\setcounter{tocdepth}{2}
\tableofcontents
\endgroup
\newpage

\section{Introduction}
Causal mediation analysis studies how a treatment affects an outcome by separating pathways through intermediate variables, or mediators, from pathways through other processes \citep{baron1986moderator,pearl2022direct,imai2010identification}. This decomposition is useful in biomedical and public health research, where an intervention may influence a clinical outcome through a biological or behavioral mechanism \citep{albert2008mediation,richiardi2013mediation}. Within the potential-outcomes framework \citep{rubin1974estimating}, natural direct and indirect effects are defined by contrasts of counterfactual means, which causal assumptions relate to observed data \citep{imai2010identification}. Estimating an indirect effect characterizes a pathway, while statistical inference quantifies the precision of that estimate and supports comparisons across studies or interventions. Semiparametric influence-function-based methods provide an inferential construction by combining nuisance functions for treatment assignment, the mediator distribution, and the outcome response \citep{tchetgen2012semiparametric,farbmacher2022causal}. Conventional implementations fit these potentially nonlinear functions separately in each study; uncertainty in the resulting effect estimate depends on how their estimation errors propagate through the estimator.

Tabular foundation models offer a way to reuse nuisance learning across studies. TabPFN and TabICL learn from collections of datasets and predict on a new table using labeled observations as context \citep{hollmann2022tabpfn,hollmann2025accurate,qu2025tabicl}, while CausalPFN, CausalFM, and Do-PFN extend this approach to treatment-effect or interventional prediction \citep{balazadeh2026causalpfn,ma2026foundation,robertson2026pfn}. Fluctuation-supervised pretraining constructs an influence-function teacher and variance target from simulator-known nuisance functions \citep{zhang2026learning}. For frequentist treatment-effect inference, \citet{melnychuk2026frequentist} use nuisance posteriors in a one-step posterior correction, whereas TabCon learns regression-function estimation errors using specialist heads and a post-trained router \citep{ye2026can}. In recent work, TabCUQ uses outcome-perturbed contexts to assess context variation and limited-context bias for frozen tabular predictors \citep{shi2026tabcuq}. These methods advance reusable point   estimation  and statistical inference through different sources of supervision and uncertainty constructions.   Their uncertainty modules use oracle nuisance targets, posterior corrections, residual experts with routing, or repeated context perturbations. They leave open how to combine adaptation of a frozen predictor from observed labels with a separately post-trained error-quantile head, so nuisance learning and interval construction can be reused without oracle nuisance supervision or repeated resampling.

We propose \textbf{MediPFN}, a framework for reusable nuisance learning and statistical inference for the \emph{natural indirect effect} (NIE). MediPFN adapts a pretrained tabular foundation model using observed treatment and outcome labels from simulated tables. On a new dataset, the adapted predictor supplies the conditional relationships needed by an influence-function-based estimator.   To construct intervals, a  separate error-quantile head learns  quantiles of  the estimator's   standardized error  across tables with known effects   and is refined through supervised post-training. Its quantiles then yield confidence intervals  without repeated bootstrap refitting.

Our contributions are   fourfold. \textbf{(i)} We develop a framework for amortized statistical inference in causal mediation analysis, combining reusable nuisance prediction with an influence-function-based estimator. \textbf{(ii)} We design an architecture that couples a trainable adaptation head with a frozen tabular backbone and a separate error-quantile head, so nuisance learning and uncertainty quantification use supervision appropriate to their respective tasks. \textbf{(iii)} We   develop a training and post-training procedure that fits the error-quantile  head on simulated tables with known indirect effects,   refines  it through supervised fine-tuning (SFT), and  calibrates the resulting intervals. We  derive coverage bounds for   this construction. \textbf{(iv)} We evaluate MediPFN across synthetic, distribution-shift, stress-test,  and semi-synthetic   settings, comparing  coverage, interval length, and computational cost. Under matched timing conditions, MediPFN is  59.1   times faster than bootstrap, measured by the median of paired per-table ratios.

\subsection{Related Work}
\noindent\textbf{Causal mediation analysis.} The potential-outcomes framework defines direct and indirect effects through potential mediator and outcome variables and identifies them under assumptions such as consistency, positivity, and absence of unmeasured confounding \citep{pearl2022direct,imai2010identification}. Building on this formulation, \citet{tchetgen2012semiparametric} derive efficient influence functions and multiply robust estimators of mediation functionals, while \citet{farbmacher2022causal} combine cross-fitting with flexible machine-learning methods for nuisance estimation. SIGMA combines causal discovery, influence functions, and Bayesian model averaging to account for uncertainty over mediation pathways \citep{li2026decoding}. These approaches provide identification and inference tools for mediation analysis, with conventional implementations fitting nuisance models separately for each dataset. MediPFN reuses nuisance learning across datasets and learns the resulting estimator's error distribution in advance, reducing the repeated fitting required for uncertainty quantification.

\smallskip
\noindent\textbf{Tabular and causal foundation models.} TabPFN and TabICL train models for in-context tabular prediction \citep{hollmann2022tabpfn,hollmann2025accurate,qu2025tabicl}, and TabDPT incorporates real tables in pretraining \citep{ma2026tabdpt}. Extensions to causal inference adapt this reuse of information to different identification settings: Do-PFN predicts interventional outcomes from observational tables \citep{robertson2026pfn}, CausalPFN amortizes treatment-effect estimation and posterior uncertainty \citep{balazadeh2026causalpfn}, and CausalFM constructs structural priors for several causal identification settings \citep{ma2026foundation}. ProximalFM uses proxy data for hidden confounding and learns a posterior for conditional treatment effects from Monte Carlo counterfactual targets generated by proximal causal models \citep{muller2026proximalfm}. MediPFN adapts a pretrained tabular predictor to the conditional relationships needed for mediation analysis using observed treatment and outcome labels; known indirect effects from simulated tables supervise a separate uncertainty head.

\smallskip
\noindent\textbf{Uncertainty quantification and calibration.} Confidence interval construction is a classical problem in statistical inference. Wald intervals combine an estimated sampling scale with a normal approximation, while bootstrap intervals approximate sampling variation through repeated resampling. More recently, fluctuation-supervised pretraining uses simulator-known nuisance functions to construct influence-function targets and supervise a variance head \citep{zhang2026learning}, and \citet{melnychuk2026frequentist} use nuisance posteriors in a one-step posterior correction for frequentist treatment-effect inference. Conformalized quantile regression calibrates prediction intervals using held-out outcomes \citep{romano2019conformalized}, with weighted conformal methods addressing covariate shift under additional assumptions \citep{tibshirani2019conformal}. For frozen tabular predictors, TabCUQ combines dispersion from outcome-perturbed contexts with a maximum-deviation correction for limited-context bias \citep{shi2026tabcuq}. TabCon learns regression-function estimation errors through residual experts and a post-trained router \citep{ye2026can}. MediPFN learns quantiles of an indirect-effect estimator's standardized error from simulated tables, then fine-tunes the error head and independently calibrates its intervals. Its nuisance predictor learns from observed labels without oracle nuisance targets, and its confidence intervals describe uncertainty in an estimated causal effect.

\section{Causal Target and Identification}\label{sec:identification}
Let $O=(Y,A,X,M)$ denote an observation from the target population, where $A\in\{0,1\}$ is a binary treatment, $X\in\mathbb R^p$ contains $p$ pretreatment covariates, $M$ is a mediator, and $Y$ is an outcome. We observe the table $D_n=(O_1,\ldots,O_n)$ of $n$ independent rows from this population and use $\ind(E)$ for the indicator of an event $E$.

For each treatment value $a$, let $M^{(a)}$ denote the potential mediator and let $Y^{(a,m)}$ denote the potential outcome when the treatment and mediator are set to $a$ and $m$, respectively. We use the shorthand $Y^{(a)}=Y^{(a,M^{(a)})}$ to decompose the total effect as
\begin{equation}\label{eq:te-decomposition}
\underbrace{\E[Y^{(1)}-Y^{(0)}]}_{\text{total effect}}
=\underbrace{\E[Y^{(1,M^{(1)})}-Y^{(1,M^{(0)})}]}_{\text{natural indirect effect}}
+\underbrace{\E[Y^{(1,M^{(0)})}-Y^{(0,M^{(0)})}]}_{\text{natural direct effect}}.
\end{equation}
For $a,a'\in\{0,1\}$, define $\theta_{aa'}=\E[Y^{(a,M^{(a')})}]$. The corresponding natural indirect, natural direct, and total effects are $\theta_{\mathrm{NIE}}=\theta_{11}-\theta_{10}$, $\nu=\theta_{10}-\theta_{00}$, and $\tau=\theta_{11}-\theta_{00}=\theta_{\mathrm{NIE}}+\nu$, respectively. We focus on $\theta_{\mathrm{NIE}}$, whose identification requires the cross-world mean $\theta_{10}$ since this mean combines the outcome under treatment with the mediator under control. We therefore impose the following standard causal assumptions.
\begin{assumption}[Consistency]\label{ass:consistency}
For each $a\in\{0,1\}$ and mediator value $m$, $A=a$ implies $M=M^{(a)}$, and $A=a,M=m$ implies $Y=Y^{(a,m)}$.
\end{assumption}
\begin{assumption}[Positivity]\label{ass:positivity}
For every $x$ in the support of $X$, $0<\Pr(A=1\mid X=x)<1$. Moreover, for each $a\in\{0,1\}$ and each $m$ in the support of $M$ given $X=x$, the conditional mediator mass or density, denoted by $p(m\mid A=a,X=x)$, is positive.
\end{assumption}
\begin{assumption}[Sequential exchangeability]\label{ass:sequential}
For all $a,a'\in\{0,1\}$ and mediator values $m$,
\begin{equation}\label{eq:exchangeability}
\{Y^{(a,m)},M^{(a')}\}\perp\!\!\!\perp A\mid X,
\qquad
Y^{(a,m)}\perp\!\!\!\perp M^{(a')}\mid A=a',X.
\end{equation}
\end{assumption}
Assumption~\ref{ass:consistency} links the observed mediator and outcome to their corresponding potential values. Assumption~\ref{ass:positivity} ensures overlap in treatment assignment and mediator distributions given $X$. The first part of Assumption~\ref{ass:sequential} rules out unmeasured confounding of the treatment--outcome and treatment--mediator relations given $X$. The second part requires conditional independence between potential outcomes and potential mediators, including those under different treatment assignments. This condition excludes an unmeasured mediator--outcome confounder induced by treatment.

Under Assumptions~\ref{ass:consistency}--\ref{ass:sequential}, the mediation formula identifies the cross-world mean \citep{pearl2022direct,imai2010identification}:
\begin{equation}\label{eq:mediation-formula}
\theta_{10}=\E_X\!\left[\int \E[Y\mid A=1,X,M=m]\,dP(m\mid A=0,X)\right]
=\E\!\left[\E\left[\E[Y\mid A=1,X,M]\mid A=0,X\right]\right].
\end{equation}
A plug-in estimator could be constructed by estimating the conditional expectations in \eqref{eq:mediation-formula}, but it can be sensitive to regression-model misspecification. We instead use an efficient influence-function-based estimator of $\theta_{10}$ \citep{tchetgen2012semiparametric}. We define the following nuisance functions:
  \begin{align}\label{eq:nuisance}
\pi_1(x)&=\Pr(A=1\mid X=x), &
\mu_1(x,m)&=\E[Y\mid A=1,X=x,M=m],\\
\pi_2(x,m)&=\frac{p(m\mid A=0,X=x)}{p(m\mid A=1,X=x)}, &
\mu_{10}(x)&=\E[\mu_1(X,M)\mid A=0,X=x].\nonumber
\end{align}
  Here $\pi_1(x)$ is the propensity score. To estimate $\pi_2$, we introduce the auxiliary propensity score $\rho(x,m)=\Pr(A=1\mid X=x,M=m)$.  Bayes' rule   gives
\begin{equation}\label{eq:bayes-nuisance}
\pi_2(x,m)
=\frac{1-\rho(x,m)}{\rho(x,m)}\frac{\pi_1(x)}{1-\pi_1(x)}.
\end{equation}
  The  efficient influence function  of $\theta_{10}$  is
\begin{align}\label{eq:eif-theta10}
\psi_{10}(O)={}&\mu_{10}(X)-\theta_{10}
+\frac{A\pi_2(X,M)}{\pi_1(X)}\{Y-\mu_1(X,M)\}\nonumber\\
&+\frac{1-A}{1-\pi_1(X)}\{\mu_1(X,M)-\mu_{10}(X)\}.
\end{align}
Under   standard  conditions on the nuisance functions, the resulting estimator has multiple-robustness properties \citep{tchetgen2012semiparametric}.

To obtain the NIE and complete the effect decomposition, we also identify the two same-treatment means. Under Assumptions~\ref{ass:consistency}--\ref{ass:sequential},
\begin{equation}\label{eq:identify}
\theta_{11}=\E[Y^{(1)}]=\E\!\left[\E[Y^{(1)}\mid X]\right]=\E\!\left[\E[Y^{(1)}\mid A=1,X]\right]=\E\!\left[\E[Y\mid A=1,X]\right].
\end{equation}
Likewise,
\begin{equation}\label{eq:identify-theta00}
\theta_{00}=\E[Y^{(0)}]=\E\!\left[\E[Y^{(0)}\mid X]\right]=\E\!\left[\E[Y^{(0)}\mid A=0,X]\right]=\E\!\left[\E[Y\mid A=0,X]\right].
\end{equation}
These identified means yield $\theta_{\mathrm{NIE}}=\theta_{11}-\theta_{10}$ and $\nu=\theta_{10}-\theta_{00}$. Let $\mu_{11}(x)=\E[Y\mid A=1,X=x]$. The efficient influence function of $\theta_{11}$ is
\begin{equation}\label{eq:eif-theta11}
\psi_{11}(O)=\frac{A}{\pi_1(X)}\{Y-\mu_{11}(X)\}+\mu_{11}(X)-\theta_{11}.
\end{equation}
The augmented inverse-probability-weighted (AIPW) estimator uses the empirical analogue of this influence function \citep{bang2005doubly}.

\section{Model Training}
In this section, we develop   MediPFN, a  framework for estimating natural indirect effects and constructing confidence intervals. We first define the influence-function-based estimator, then describe nuisance-function adaptation and error-distribution learning for inference.

\subsection{Influence-Function-Based Estimation}\label{sec:if-estimation}
Cross-fitting estimates nuisance functions on one part of the data and evaluates the influence function on a held-out part \citep{chernozhukov2018double}. Partition the observations into $K$ folds. For each fold, the adapted predictor uses the other $K-1$ folds as labeled context and estimates the required functions on the held-out set. Repeating this procedure across all folds supplies predictions for every observation, whose contributions are then averaged. The model parameters remain fixed while the observed context changes. We suppress the fold index on the fitted functions below, with each prediction computed from the complementary folds.

The resulting estimate of $\theta_{11}$ is
\begin{equation}\label{eq:aipw-theta11}
\widehat\theta_{11}
=\frac{1}{n}\sum_{i=1}^n\left[\widehat\mu_{11}(X_i)
+\frac{A_i}{\widehat\pi_1(X_i)}\{Y_i-\widehat\mu_{11}(X_i)\}\right].
\end{equation}
For the cross-world mean $\theta_{10}$, replacing the nuisance functions in \eqref{eq:eif-theta10} by their fitted estimates gives
\begin{align}\label{eq:if-estimator-theta0}
\widehat\theta_{10}=\frac{1}{n}\sum_{i=1}^{n}\bigg[&\widehat\mu_{10}(X_i)
+\frac{A_i\widehat\pi_2(X_i,M_i)}{\widehat\pi_1(X_i)}\{Y_i-\widehat\mu_1(X_i,M_i)\}\nonumber\\
&+\frac{1-A_i}{1-\widehat\pi_1(X_i)}\{\widehat\mu_1(X_i,M_i)-\widehat\mu_{10}(X_i)\}\bigg].
\end{align}
Combining the two estimates gives the natural indirect effect estimator
\begin{equation}\label{eq:score}
\widehat\theta_{\mathrm{NIE}}=\widehat\theta_{11}-\widehat\theta_{10}.
\end{equation}
Within each set of context rows, an additional split produces out-of-fold outcome predictions for the control-side regression $\widehat\mu_{10}$; $\widehat\mu_{11}$ is fitted directly from treated outcomes. Let $\widehat\phi_{11,i}$ and $\widehat\phi_{10,i}$ denote the uncentered summands in the estimators of $\theta_{11}$ and $\theta_{10}$. \thisrevision{We estimate the standard deviation of $\widehat\theta_{\mathrm{NIE}}$ by $\widehat s$, the empirical standard deviation of $\widehat\phi_{11,i}-\widehat\phi_{10,i}$ across rows divided by $\sqrt n$.} The Wald interval at nominal level $1-\alpha$ has radius $z_{1-\alpha/2}\widehat s$.

\subsection{MediPFN Architecture and Training}
\begin{figure}[t]
\centering
\includegraphics[width=\linewidth]{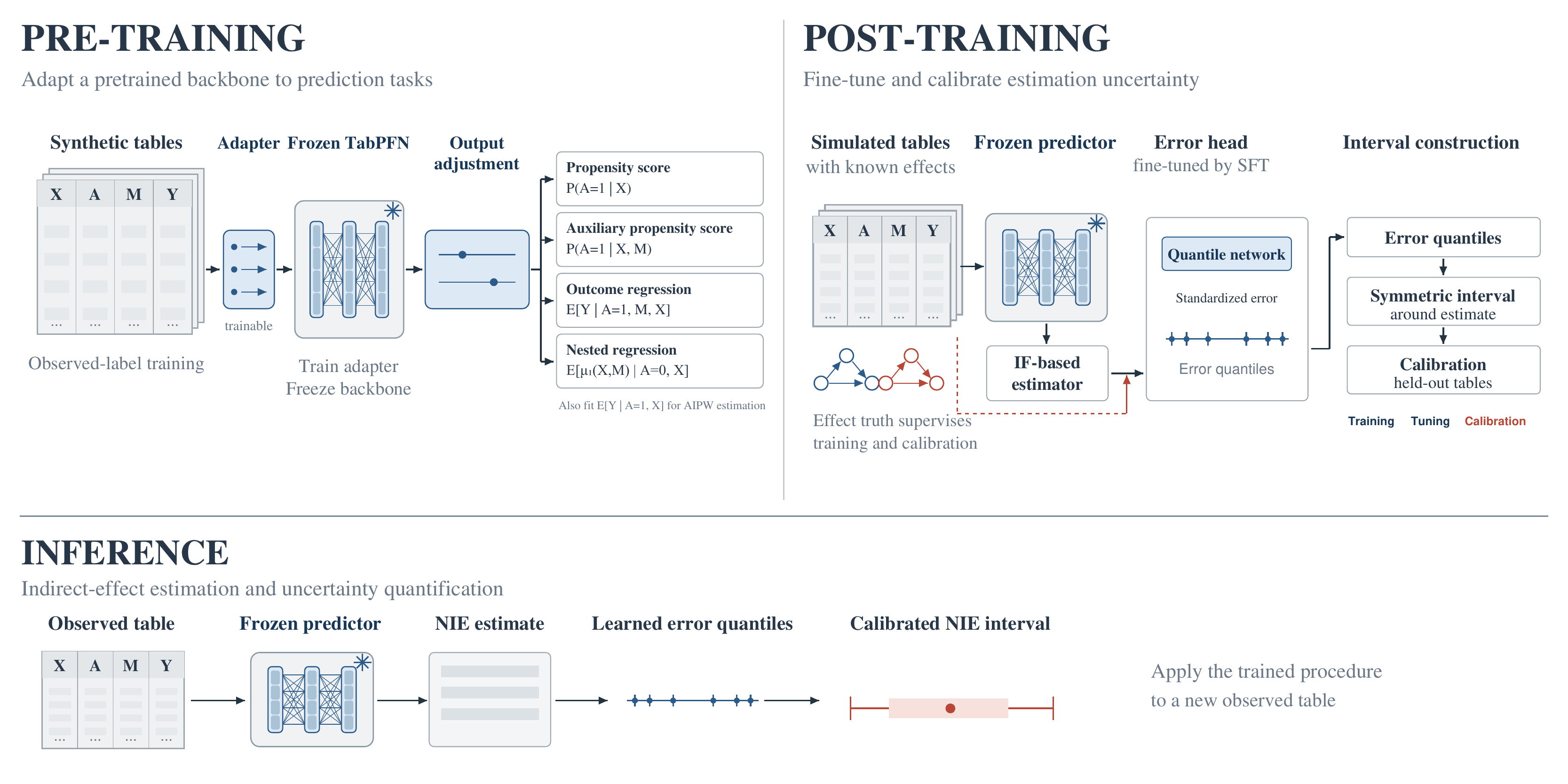}
\caption{A schematic overview of MediPFN adaptation, uncertainty training, and inference.}
\label{fig:pipeline}
\end{figure}

\paragraph{Overview.}   To construct confidence  intervals for the natural indirect effect, we  account for errors introduced by estimating the conditional means and propensity scores in \eqref{eq:score}. MediPFN first adapts \currentchange{TabPFN V3 (\citealt{grinsztajn2026tabpfn})} to these regression and classification problems, then learns the distribution of the resulting estimator's error from synthetic tables (datasets generated from specified mechanisms with known indirect effects). The learned error quantiles, symmetrization, and independent calibration determine the final confidence interval. Figure~\ref{fig:pipeline} summarizes training and inference.

\paragraph{Adaptation head.} Nuisance estimation involves regression and classification problems with different sets of input variables. We   place a shared input adapter before TabPFN  and a role-specific   output adjustment after it, while keeping the TabPFN weights fixed \citep{hollmann2022tabpfn,hollmann2025accurate}. For each standardized feature value $z$, the transformation uses $z/5$, $(z/5)^2$, an indicator distinguishing classification from regression, and a constant. The first two inputs permit linear and nonlinear adjustments to feature magnitude; the indicator allows the transformation to depend on the prediction problem, and the constant supplies an offset.   The head learns a small additive adjustment to each input value  $z$, and the adjusted value is passed to TabPFN.   The output adjustment has separate  scale and offset   parameters for classification and regression. For classification, we apply the adjustment on the log-odds scale and map the result back to a propensity score. For an outcome prediction, it rescales and shifts the predicted mean.

\begin{remark}
Parameter sharing lets us use the same input adapter for prediction from $X$, $(X,M)$, and $(X,A,M)$ across tables with different numbers of covariates or mediators. TabPFN then models interactions among the transformed features. TFM-Retouche also adapts the inputs to a frozen backbone, using a joint residual transformation of the feature vector fitted separately to each downstream dataset (\citealt{nguyen2026tfm}).
\end{remark}

We train   the input transformation and output adjustment jointly  across synthetic tables, keeping all TabPFN weights fixed. Each table is divided into   labeled context rows $S$  and held-out rows   for evaluating predictions.
The fitted predictions yield the four nuisance functions as follows:
\begin{enumerate}
\renewcommand{\labelenumi}{(\roman{enumi})}\setlength{\itemsep}{2pt}\setlength{\parsep}{0pt}
\item \underline{\emph{Propensity scores.}} Predicting treatment from $X$ yields $\widehat\pi_1$, while predicting treatment from $(X,M)$ yields $\widehat\rho$. We use these fitted   values  in \eqref{eq:bayes-nuisance} to obtain $\widehat\pi_2$.
\item \underline{\emph{Outcome and nested regressions.}} Regressing $Y$ on $(X,A,M)$ and evaluating at $A=1$ yields $\widehat\mu_1$. Within the context rows, an additional split provides held-out predictions of this regression.   These predictions serve as context labels for the regression  on $X$ among control rows, yielding  $\widehat\mu_{10}$  with the adaptation parameters fixed. A separate regression of $Y$ on $X$ among treated context rows gives $\widehat\mu_{11}$ for $\theta_{11}$.
\end{enumerate}
At inference, the fitted parameters remain fixed and each new table supplies its own context rows. Appendix~\ref{sec:network-details} provides the construction and training details.

\paragraph{Error distribution and SFT.}\label{sec:confidence-scaling} We   next  evaluate the fixed estimator on synthetic tables with known effect $\theta_{\mathrm{NIE}}$ and calculate its standardized error $Z=(\widehat\theta_{\mathrm{NIE}}-\theta_{\mathrm{NIE}})/\widehat s$. Let $T=T(D_n)$ collect observable summaries of the table, fitted  values, and estimator, together with pooled backbone representations. A neural network predicts the error quantiles from $T$, with nonnegative increments between successive outputs to preserve their ordering. At quantile level $\tau$, the training loss is $\ell_\tau(u)=u\{\tau-\ind(u<0)\}$, where $u=Z-\widehat z_\tau(T)$; we average this loss over training tables and quantile levels.

Next, SFT updates the error head using two equally weighted collections: the original synthetic tables and semi-synthetic tables that combine observed covariates with simulated treatment, mediator, and outcome values. The latter construction also supplies a known indirect effect. Sampling balances mechanism groups within the original collection and covariate sources within the semi-synthetic collection. The backbone and adaptation head remain fixed. Separate development tables are used to select the checkpoint and interval construction, after which independent calibration determines the final expansion.

For a nominal level $1-\alpha$, let $\hat z_\ell(T)$ and $\hat z_u(T)$ denote predicted error quantiles at levels $\alpha/2$ and $1-\alpha/2$. Their direct inversion defines the initial interval (the subscript $0$ denotes the interval before calibration)
\begin{equation}\label{eq:ci}
C_{\alpha,0}(D_n)=[\widehat L_\alpha,\widehat U_\alpha],\quad
\widehat L_\alpha=\hat\theta_{\mathrm{NIE}}-\hat s\hat z_u,\quad
\widehat U_\alpha=\hat\theta_{\mathrm{NIE}}-\hat s\hat z_\ell.
\end{equation}
For the final interval, let $r_\alpha(T)=\max(|\hat z_\ell(T)|,|\hat z_u(T)|)$. Symmetrization gives $[\hat\theta_{\mathrm{NIE}}\pm\hat s r_\alpha(T)]$, which contains \eqref{eq:ci}.

\paragraph{Independent calibration.} Let $\mathcal C_m=\{(D_i^{\mathrm{cal}},\theta_{\mathrm{NIE},i}^{\mathrm{cal}})\}_{i=1}^m$ denote the independent calibration tasks and their known effects, where each $D_i^{\mathrm{cal}}$ includes its table and realized folds. For calibration mechanism $i$, let $Z_i$ be its standardized error and $T_i$ its observable features. The nonnegative score
\begin{equation}\label{eq:cal-score}
 S_i=\max(0,|Z_i|-r_\alpha(T_i))
\end{equation}
measures the amount by which the error exceeds the initial radius. We set $c_\alpha^0$ to the $\lceil(m+1)(1-\alpha)\rceil$th ordered score from the independent calibration mechanisms. To preserve nested intervals across the reported nominal levels, we set $c_\alpha$ to the largest $c_\beta^0$ among the current and lower nominal coverage levels. Hence the resulting radius and confidence interval are
\begin{equation}\label{eq:ci-calibrated}
\begin{aligned}
R_\alpha(D_n;\mathcal C_m)&=\hat s\{r_\alpha(T)+c_\alpha\},\\
C_\alpha(D_n;\mathcal C_m)&=\left[\hat\theta_{\mathrm{NIE}}-R_\alpha(D_n;\mathcal C_m),\;\hat\theta_{\mathrm{NIE}}+R_\alpha(D_n;\mathcal C_m)\right].
\end{aligned}
\end{equation}
The observed table and its folds determine $\hat\theta_{\mathrm{NIE}}$, $\hat s$, and $T$; for notational convenience, we suppress the folds. The independent calibration collection determines $c_\alpha=c_\alpha(\mathcal C_m)$ for the fixed model. Training, development, calibration, and testing use distinct mechanism sets.

\paragraph{Inference.} On a new observed table, $K$-fold nuisance predictions supply $\hat\theta_{\mathrm{NIE}}$, $\hat s$, and the observable features $T$. The fixed error head and calibration correction then give \eqref{eq:ci-calibrated}.

\subsection{Coverage Bounds}
We now describe how error-quantile accuracy and independent calibration determine coverage. Throughout this section, $\mathbb P$ and $\E$ denote probability and expectation conditional on the model, normalization, and development selection fixed before calibration. They average over the target mechanism, its sampled table and randomized folds, and, when present, the independent calibration sample. \currentchange{Let $B$ indicate that the procedure returns an interval and set $\delta=\mathbb P(B=0)$.} On successful tasks, define $F(z\mid T)=\mathbb P(Z\leq z\mid T,B=1)$. Let $\mathcal R$ denote the expected mean excess pinball loss of the two predicted quantiles over their true conditional quantiles, conditional on $B=1$.

  \begin{assumption}[Error-distribution regularity] \label{ass:coverage-conditions}
 \begin{enumerate}
\renewcommand{\labelenumi}{(\roman{enumi})}
\item The success probability satisfies $\mathbb P(B=1)>0$. On successful tasks, $\hat s>0$, $F(\cdot\mid T)$ is continuous, and $\hat z_\ell(T)\leq\hat z_u(T)$.
\item On successful tasks, the conditional density of $Z$ is bounded above by   $C$  and below by   $c>0$  between each predicted quantile and its true counterpart. Moreover, $\E[|Z|\mid T,B=1]<\infty$ almost surely and $\mathcal R<\infty$.
 \end{enumerate}
\end{assumption}
 Assumption~\ref{ass:coverage-conditions} comprises mild local regularity conditions for continuous error distributions with uniformly bounded, positive density over the quantile ranges under study.  Part (i)   ensures that  conditional error quantiles   can be evaluated on successful tasks. Part  (ii)   requires finite loss and controls the density between each predicted and true quantile, allowing the excess pinball loss to bound quantile error.

Define the conditional tail-probability discrepancy
 \begin{equation}\label{eq:quantile-residual}
\zeta(T)=\big|F(\hat z_\ell\mid T)-\alpha/2\big|+
\big|F(\hat z_u\mid T)-(1-\alpha/2)\big|.
\end{equation}
The next proposition relates coverage of the direct interval to   this discrepancy and \currentchange{$\delta$}.
\begin{proposition}[Coverage approximation]\label{prop:coverage}
  Under Assumption~\ref{ass:coverage-conditions}(i), the  direct interval in \eqref{eq:ci} satisfies
\begin{equation}\label{eq:coverage-bound}
\left|\mathbb P\{\theta_{\mathrm{NIE}}\in C_{\alpha,0}(D_n)\}-(1-\alpha)\right|
\leq \E[\zeta(T)\ind(B=1)]+(1-\alpha)\delta.
\end{equation}
The symmetric initial interval consequently has coverage at least $1-\alpha-\E[\zeta(T)\ind(B=1)]-(1-\alpha)\delta$.
\end{proposition}

Proposition~\ref{prop:coverage} bounds the deviation of the direct interval's coverage from its nominal level by two terms: the average conditional tail-probability discrepancy on successful tasks and the probability \currentchange{$\delta$}.   Since  the symmetric interval contains the direct interval on successful tasks, the same terms also yield its coverage lower bound.

We next relate   quantile accuracy  to the excess pinball loss of the error head.
 \begin{corollary}[Conditional quantile-loss rate]\label{cor:proper-rate}
Under   Assumption~\ref{ass:coverage-conditions}(i)--(ii),
  \begin{equation}\label{eq:risk-bound}
\mathbb P\{\theta_{\mathrm{NIE}}\in[\hat\theta_{\mathrm{NIE}}\pm\hat s r_\alpha(T)]\}
\geq (1-\alpha)(1-\delta)-2C\sqrt{2\mathcal R/c}.
\end{equation}
  For  $N$ training   tables, a target-distribution excess-risk rate $\mathcal R=O(N^{-\beta})$ with $\beta>0$ yields an approximation contribution of order  $O(N^{-\beta/2})$.
\end{corollary}
Corollary~\ref{cor:proper-rate}   \currentchange{connects coverage of the symmetric interval to $\delta$ and the error head's excess quantile loss.} Under the stated excess-risk rate,  the  quantile approximation term decreases as $N^{-\beta/2}$. This rate statement is conditional on control of the target-distribution excess risk.

For the calibrated interval, let  $S_1,\ldots,S_m$ be the scores in \eqref{eq:cal-score}, and define $S_*$ analogously for an independent target task, \currentchange{with $S_*=+\infty$ if $B=0$}. Let $k=\lceil(m+1)(1-\alpha)\rceil$ and let  $c_\alpha^0$ be the $k$th ordered calibration score. Define  the calibration-to-target mismatch
\begin{equation}\label{eq:calibration-shift}
\Delta_m=\sup_{t\in\mathbb R}\left|m^{-1}\sum_{i=1}^m\mathbb P(S_i\leq t)-\mathbb P(S_*\leq t)\right|.
\end{equation}
  \begin{assumption}[Calibration sampling]\label{ass:calibration-sampling}
The calibration scores are independent and almost surely finite, the target task is independent of them, and $k\leq m$.
\end{assumption}
Assumption~\ref{ass:calibration-sampling} separates calibration from target-task sampling while allowing the calibration mechanisms to  have different score distributions.  The mismatch $\Delta_m$ records how these distributions differ from the target distribution.

The next theorem characterizes coverage of the calibrated interval under heterogeneous mechanisms.
 \begin{theorem}[Calibrated coverage under heterogeneous mechanisms]\label{thm:calibrated-coverage}
Under Assumption~  \ref{ass:calibration-sampling}, the interval in \eqref{eq:ci-calibrated} satisfies
\begin{equation}\label{eq:calibration-bound}
\mathbb P\{\theta_{\mathrm{NIE},*}\in C_\alpha(D_*;\mathcal C_m)\}\geq
\max\left(0,\frac{k}{m}-\Delta_m-\sqrt{\frac{\pi}{8m}}\right).
\end{equation}
 \end{theorem}
Theorem~\ref{thm:calibrated-coverage} gives a coverage bound for independent calibration mechanisms with potentially different distributions. In the bound, $k/m$ is determined by the calibration rank, while $\Delta_m$ measures the discrepancy between the average calibration-score distribution and the target-score distribution. The term $\sqrt{\pi/(8m)}$ is a finite-calibration-sample contribution. For a fixed nominal level, $k/m$ tends to $1-\alpha$ as $m$ increases, so the bound approaches $1-\alpha$   when  $\Delta_m$ also tends to zero. In the exchangeable case, the rank argument  gives marginal coverage of at least $k/(m+1)\geq 1-\alpha$.  On a successful task, the interval length is
\[
\operatorname{length}(C_\alpha(D_n;\mathcal C_m))=2R_\alpha(D_n;\mathcal C_m).
\]
  The proofs of Proposition~\ref{prop:coverage}, Corollary~\ref{cor:proper-rate}, and Theorem~\ref{thm:calibrated-coverage} are given in Appendix~\ref{sec:coverage-proofs}.

\section{Experiments}
\currentchange{In this section, we evaluate MediPFN on synthetic data and data with real covariates. We compare coverage, interval length, point estimation accuracy, and inference time. We also vary sample size, outcome noise, and the data-generating mechanism in stress tests and examine the contribution of fine-tuning and interval construction in an ablation study.}
\subsection{Evaluation Design}\label{sec:evaluation-design}
\noindent\roundchange{\textbf{Synthetic data.} The main synthetic evaluation uses 1,152 tables with $n\in\{200,600,1000\}$. The 768 in-distribution (ID) tables draw new mechanisms from families used to train the adapter. The 288 source-shift tables draw from two held-out families. The 96 composition-shift tables combine structural components in a configuration held out within an adapter training family. All evaluation tables are generated independently of the training tables (Appendix~\ref{sec:synthetic-banks}).}

\smallskip
\noindent\currentchange{\textbf{Real covariates.} We use feature matrices from eight held-out OpenML datasets (\citealt{vanschoren2014openml}). The main comparison uses cnae-9, adult, eucalyptus, and car. A second set uses pc1, MagicTelescope, climate-model-simulation-crashes, and ilpd to evaluate the selected procedure. We resample covariate rows with replacement to generate 128 tables per source (Appendix~\ref{sec:supp-data}).}

\smallskip
\noindent\roundchange{\textbf{Stress tests and reference benchmark.} A separate set of 576 stress tables varies sample size, outcome noise, and mechanism shift. The pooled ablation uses these tables to compare the stages of interval construction. \currentchange{The reference benchmark comprises 2,336 tables:} 1,152 synthetic evaluation tables, 480 stress tables, 80 sample-size scaling tables, and 624 semi-synthetic tables.}

\smallskip
\noindent\roundchange{\textbf{Evaluation criteria.} We focus on $95\%$ nominal coverage ($\alpha=0.05$). Appendix~\ref{sec:additional-levels} reports results at $90\%$ ($\alpha=0.10$) and $99\%$ ($\alpha=0.01$). \thisrevision{For intervals, we report empirical coverage $\widehat p_\alpha$ as a percentage and mean length $\bar\ell_\alpha$. For point estimates, we report bias, sampling standard deviation (SD), and root mean squared error (RMSE).} See Appendix~\ref{sec:additional-evaluation-details} for more details.}

\paragraph{\underline{Comparison methods}.}\leavevmode We compare MediPFN against five methods.
\begin{enumerate}
\setlength{\itemsep}{1pt}\setlength{\parsep}{0pt}
\item \underline{\emph{Linear SEM}} regresses the mediators on treatment and covariates, and the outcome on treatment, mediators, and covariates using main effects. We construct its interval using a sandwich covariance estimate and the delta method.
\item \underline{\emph{RFF}} augments the predictors with random Fourier features to allow nonlinear nuisance functions. We use logistic regression for propensity scores and ridge regression for the conditional means, then combine the cross-fitted predictions using \eqref{eq:score} and form a Wald interval from the empirical score standard error.
\item \underline{\emph{RF}} uses random-forest classifiers and regressors for the propensity scores and conditional means. We use the same cross-fitted influence-function estimator and Wald interval as for RFF.
\item \underline{\emph{RF bootstrap}} fits random-forest regressions and estimates the indirect effect by averaging the difference between the two fitted nested means. We recompute this $g$-computation estimate on 50 nonparametric resamples and use the empirical percentiles as interval endpoints.
\item \underline{\emph{TabPFN V3}} uses the unadapted TabPFN V3 backbone for nuisance prediction (\citealt{grinsztajn2026tabpfn}). We construct its bootstrap interval by recomputing the cross-fitted influence-function estimate on 50 nonparametric resamples and taking empirical percentile limits.
\end{enumerate}
\subsection{Results}
\begin{figure}[!htbp]
\centering
\includegraphics[width=\linewidth]{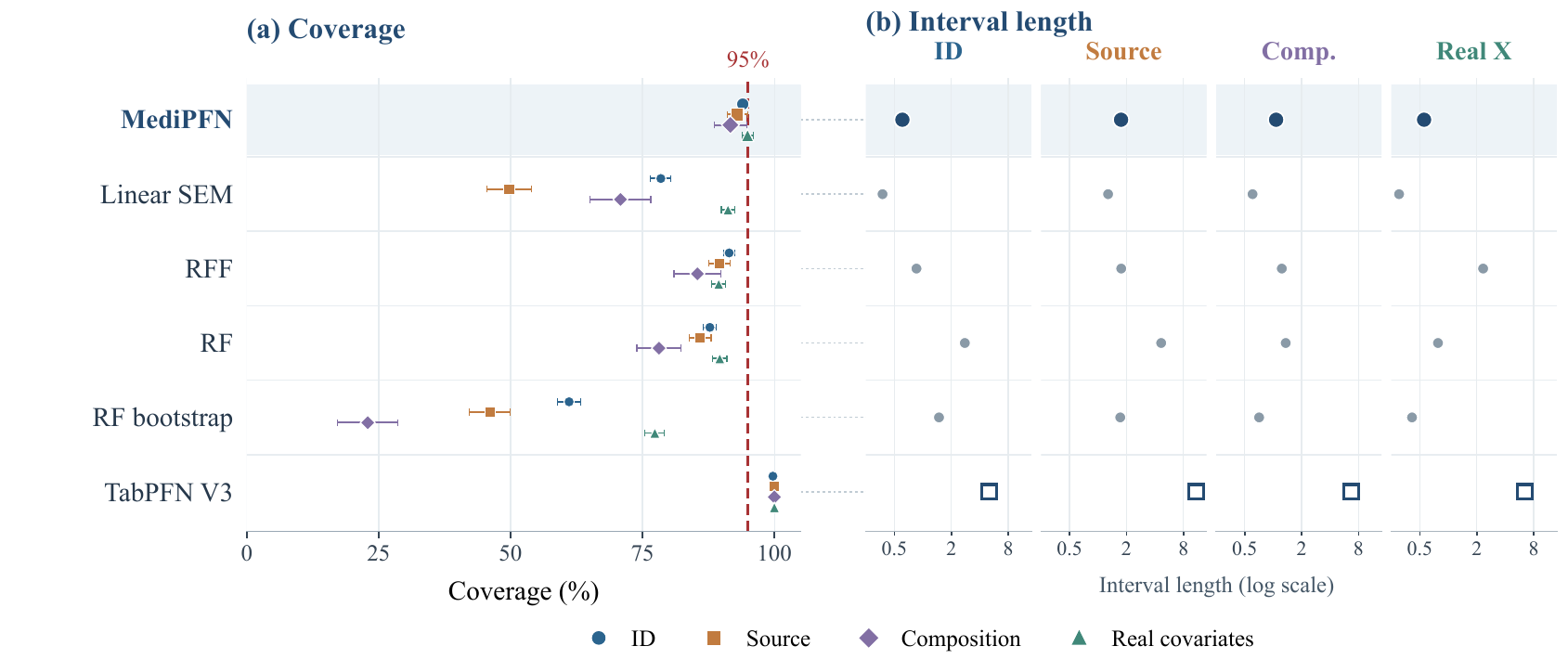}
\caption{Empirical coverage and mean interval length at the 95\% nominal level. \currentchange{Panel (a) shows coverage with mechanism-clustered standard errors. Panel (b) shows mean interval length on a logarithmic scale.}}
\label{fig:cpu-coverage}
\end{figure}

\noindent\textbf{Comparisons.}\enspace \currentchange{Table~\ref{tab:cpu-native} compares MediPFN with the five baselines on the synthetic settings and the first real-covariate set.} MediPFN attains coverage of 94.01\%, 92.97\%, and 91.67\% across the three synthetic settings. In Figure~\ref{fig:cpu-coverage}, MediPFN is closest to 95\% among the displayed methods in each setting. Its intervals are about 84\% to 91\% shorter than those of TabPFN V3, which covers at least 99.72\% in each setting. Linear SEM gives shorter intervals, with lower coverage in each setting.
\begin{table}[!htbp]
\centering\footnotesize
\setlength{\tabcolsep}{3pt}
\begin{tabular}{lcccccccc}
\toprule
Method & \multicolumn{2}{c}{ID} & \multicolumn{2}{c}{Source shift} & \multicolumn{2}{c}{Composition shift} & \multicolumn{2}{c}{Real covariates}\\
\cmidrule(lr){2-3} \cmidrule(lr){4-5} \cmidrule(lr){6-7} \cmidrule(lr){8-9}
 & $\widehat p_\alpha$ & $\bar\ell_\alpha$ & $\widehat p_\alpha$ & $\bar\ell_\alpha$ & $\widehat p_\alpha$ & $\bar\ell_\alpha$ & $\widehat p_\alpha$ & $\bar\ell_\alpha$\\
\midrule
MediPFN & \textbf{\underline{94.01}} & 0.607 & \textbf{\underline{92.97}} & 1.752 & \textbf{\underline{91.67}} & 1.074 &  \textbf{\underline{94.92}} & 0.555\\
Linear SEM & 78.48 & 0.375 & 49.74 & 1.278 & 70.83 & 0.605 & \underline{91.21} & 0.302\\
RFF  & \underline{91.43} & 0.856 & 89.58 & 1.759 & 85.42 & 1.234 & 89.45 & 2.334\\
RF  & 87.77 & 2.776 & 85.94 & 4.651 & 78.13 & 1.358 & 89.65 & 0.778\\
RF bootstrap & 61.09 & 1.481 & 46.09 & 1.718 & 22.92 & 0.710 & 77.34 & 0.413\\
TabPFN V3 & 99.72 & 5.034 & \underline{100.00} & 10.938 & \underline{100.00} & 6.643 & 100.00 & 6.439\\
\bottomrule
\end{tabular}
\caption{Interval coverage and length on the synthetic and first real-covariate evaluations. Bold underlining marks first place and underlining marks second. Coverage is ranked by closeness to 95\% and length among methods reaching 95\%.}
\label{tab:cpu-native}
\end{table}

\roundchange{Table~\ref{tab:real-sources-95} breaks down the first semi-synthetic set by OpenML dataset. On cnae-9, MediPFN ties RFF in closeness to 95\% coverage.} On cnae-9, its intervals are also the shortest among methods reaching 95\% coverage.
\begin{table}[!htbp]
\centering\footnotesize
\setlength{\tabcolsep}{3pt}
\begin{tabular}{lcccccccc}
\toprule
Method & \multicolumn{2}{c}{cnae-9} & \multicolumn{2}{c}{adult} & \multicolumn{2}{c}{eucalyptus} & \multicolumn{2}{c}{car}\\
\cmidrule(lr){2-3} \cmidrule(lr){4-5} \cmidrule(lr){6-7} \cmidrule(lr){8-9}
 & $\widehat p_\alpha$ & $\bar\ell_\alpha$ & $\widehat p_\alpha$ & $\bar\ell_\alpha$ & $\widehat p_\alpha$ & $\bar\ell_\alpha$ & $\widehat p_\alpha$ & $\bar\ell_\alpha$\\
\midrule
MediPFN & \textbf{\underline{96.88}} & \textbf{\underline{0.456}} & \textbf{\underline{94.53}} & 0.526 & \textbf{\underline{93.75}} & 0.695 & \textbf{\underline{94.53}} & 0.541\\
Linear SEM & 89.84 & 0.299 & \underline{91.41} & 0.309 & 89.84 & 0.287 & \underline{93.75} & 0.312\\
RFF  & \textbf{\underline{96.88}} & \underline{2.187} & \underline{91.41} & 3.338 & 83.59 & 2.616 & 85.94 & 1.194\\
RF  & \underline{91.41} & 1.096 & 87.50 & 0.608 & \underline{90.62} & 0.793 & 89.06 & 0.615\\
RF bootstrap & 81.25 & 0.358 & 68.75 & 0.409 & 82.81 & 0.445 & 76.56 & 0.442\\
TabPFN V3 & 100.00 & 3.337 & 100.00 & 8.224 & 100.00 & 6.120 & 100.00 & 8.076\\
\bottomrule
\end{tabular}
\caption{Interval coverage and length by real-covariate source at 95\% nominal coverage. \currentchange{Bold underlining marks first place and underlining marks second. Coverage is ranked by closeness to 95\% and length among methods reaching 95\%.} }
\label{tab:real-sources-95}
\end{table}

\medskip
\noindent\textbf{Point estimation.}\enspace Table~\ref{tab:point-variance} reports bias, root mean squared error (RMSE), and sampling standard deviation (SD), alongside interval endpoints, length, and coverage. In the ID setting, increasing the sample size from 200 to 1,000 reduces RMSE from 0.593 to 0.150 and more than halves interval length, with coverage near 95\%. Appendix~\ref{sec:additional-evaluation-details} defines the summary statistics.
\begin{table}[!htbp]
\centering\small
\setlength{\tabcolsep}{5pt}
\begin{tabular}{lcccccc}
\toprule
Setting & Bias & RMSE & SD & Interval endpoints & Interval length & Coverage (\%)\\
\midrule
ID, all & $-0.002$ & 0.369 & 0.385 & $[-0.071,\,0.536]$ & 0.607 & 94.01\\
\quad $n=200$ & $-0.019$ & 0.593 & 0.631 & $[-0.310,\,0.644]$ & 0.954 & 92.11\\
\quad $n=600$ & 0.022 & 0.182 & 0.161 & $[0.009,\,0.481]$ & 0.472 & 95.39\\
\quad $n=1000$ & $-0.008$ & 0.150 & 0.147 & $[0.088,\,0.483]$ & 0.395 & 94.52\\
Source shift & 0.076 & 1.039 & 1.211 & $[-0.097,\,1.655]$ & 1.752 & 92.97\\
Composition shift & $-0.043$ & 0.427 & 0.365 & $[-0.610,\,0.464]$ & 1.074 & 91.67\\
\bottomrule
\end{tabular}
\caption{MediPFN point estimation and interval performance on the synthetic settings.}
\label{tab:point-variance}
\end{table}

\begin{figure}[H]
\centering
\includegraphics[width=.84\linewidth]{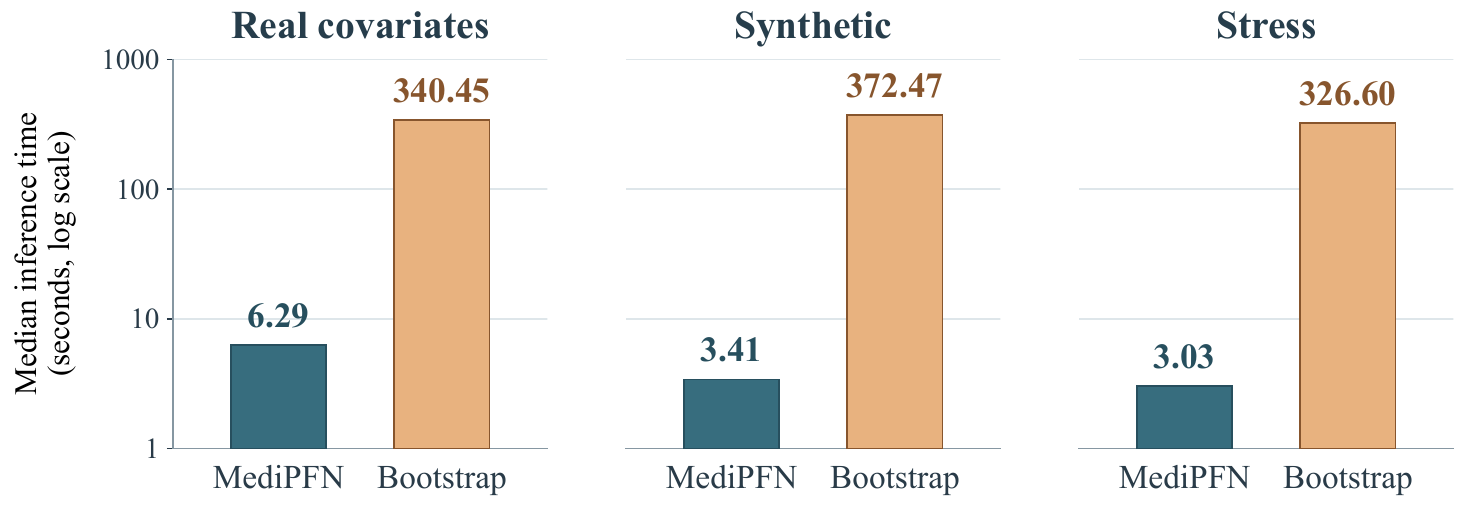}
\caption{Median inference time for MediPFN and bootstrap.}
\label{fig:cpu-timing}
\end{figure}

\medskip
\noindent\textbf{Inference time.}\enspace \roundchange{We time MediPFN and bootstrap on the same 27 real-covariate, synthetic, and stress tables.} Timing covers preprocessing, nuisance fitting, effect estimation, and interval construction (Appendix~\ref{sec:runtime-details}). Figure~\ref{fig:cpu-timing} reports group medians for bootstrap with 100 resamples. Across all 27 tables, the median of the paired bootstrap-to-MediPFN time ratios is 59.1.

\subsection{Ablation Study, Stress Tests, and Real Covariates}\label{sec:supplementary-evaluation}
In this subsection, we use paired ablations to separate the effects of fine-tuning, interval construction, and calibration. We then assess the selected procedure under changes in sample size, outcome noise, and mechanism family, and on tables generated from real covariates.

\medskip
\noindent\textbf{Ablation study.}\enspace We compare three stages of interval construction to see how fine-tuning and the final interval construction affect coverage and length. The nuisance predictor, folds, and point estimates are fixed across stages.
\begin{itemize}
\setlength{\itemsep}{1pt}\setlength{\parsep}{0pt}
\item \emph{Original head, direct.} We invert the original head's error quantiles to form the reference interval.
\item \emph{SFT, direct.} We fine-tune the error head and keep the direct interval construction.
\item \emph{SFT, symmetric and calibrated.} We center the interval at the point estimate, take the farther endpoint's distance as the radius, and apply the correction from independent calibration tasks.
\end{itemize}
For the pooled ablation, each of the three stages uses the same 576 stress tables at 95\% nominal coverage.  With direct intervals, SFT changes coverage from 84.55\% to 84.38\% and reduces mean length from 1.426 to 1.387. Appendix~\ref{sec:cpu-stress-results} reports the component-wise and semi-synthetic comparisons.

\begin{figure}[!ht]
\centering
\includegraphics[width=\linewidth]{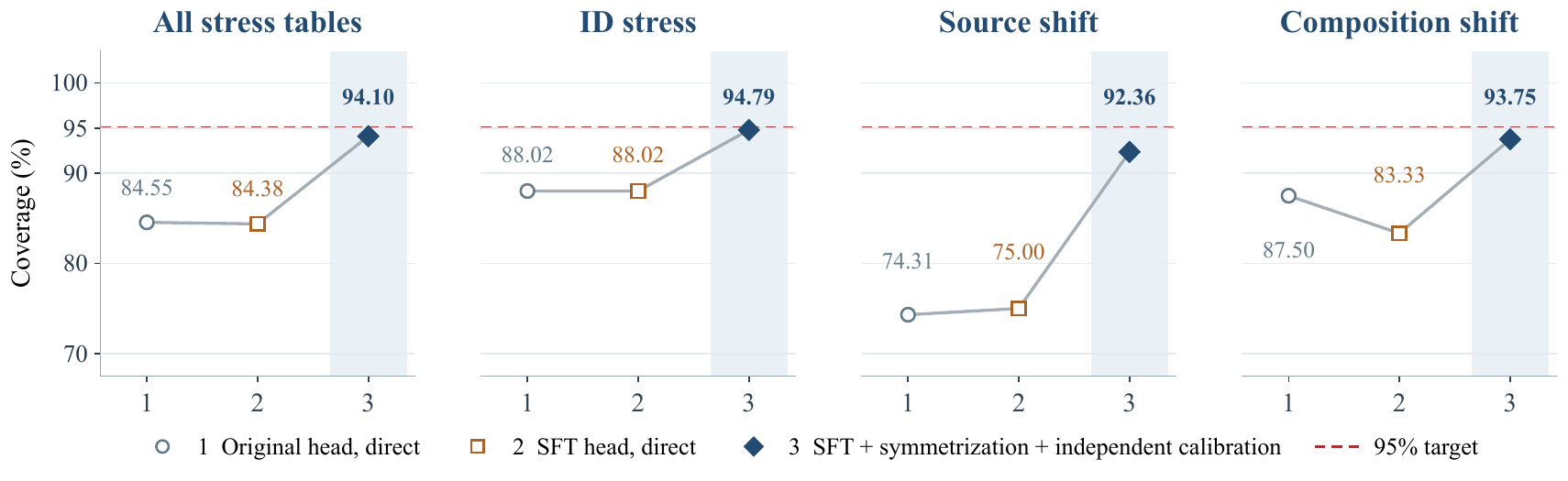}
\caption{Coverage across three stages of the interval procedure at 95\% nominal coverage.}
\label{fig:posttraining-ablation}
\end{figure}

\medskip
\noindent\textbf{Stress tests.}\enspace We examine whether coverage changes as the sample gets smaller, outcome noise becomes heavier-tailed, or the data-generating mechanism shifts. \reviewchange{We draw two tables per mechanism, for 576 tables from 288 mechanisms across 12 DGP designs.}
\begin{itemize}
\setlength{\itemsep}{1pt}\setlength{\parsep}{0pt}
\item \emph{Sample size.} We stratify the first real-covariate set by $n$. \reviewchange{The stress evaluation uses} $n\in\{64,128,256,600\}$, with 144 tables at each size.
\item \emph{Outcome noise.} \reviewchange{We add Gaussian, Student-$t_3$, or Student-$t_5$ noise to the outcomes, with 192 tables per noise type.}
\item \emph{Mechanism shift.} We use 384 tables from adaptation families, 144 from held-out families (source shift), and 48 from a held-out composition within an adaptation family (composition shift).
\end{itemize}
\reviewchange{The adapter, error head, and calibration are fixed before generating the stress tables.}

\noindent\begin{minipage}{\linewidth}
In the first real-covariate evaluation, coverage is 88.55\% on the 166 tables with $n=200$ and 94.38\% on the 178 tables with $n=600$. \reviewchange{In the stress evaluation, coverage is 92.19\% with added $t_3$ noise and 93.75\% with added $t_5$ noise, with 192 tables per noise type. } Appendix~\ref{sec:cpu-stress-results} \reviewchange{gives the paired stress summaries.}
\end{minipage}

\medskip
\noindent\textbf{Real-covariate evaluation.}\enspace MediPFN attains coverage of \currentchange{94.14\%} and 95.90\% on the two sets (Table~\ref{tab:robust-4m}). Each set contains 512 tables; the corresponding mean interval lengths are 0.554 and 3.656. \currentchange{Appendix~\ref{sec:real-strata} reports results by effect sign and sample size.}

\subsection{Sensitivity Analysis}\label{sec:sensitivity-analysis}
Next, we complement the preceding evaluation with an exploratory sensitivity analysis of the NIE interval.
An unmeasured common cause of the mediator and outcome can bias estimates of the NIE even when treatment is assigned independently of the potential variables given $X$. We assess how large this bias would have to be to change the conclusion drawn from the MediPFN interval. We consider a setting with a binary pretreatment variable $U$ \citep{vanderweele2010bias}. We retain consistency and positivity and assume that $A$ is independent of $U$ and the potential variables conditional on $X$. We replace the exchangeability condition in Assumption~\ref{ass:sequential} with $Y^{(1,m)}\perp\!\!\!\perp M^{(a)}\mid (X,U)$ for $a=0,1$.

Let $\theta_{\mathrm{obs}}$ denote the observed-data contrast
\[
\theta_{\mathrm{obs}}=\E_X\!\left[\E[Y\mid A=1,X]-\int \E[Y\mid A=1,M=m,X]\,dP(m\mid A=0,X)\right].
\]
For each $(m,x)$, define the outcome and prevalence contrasts
\begin{align*}
\gamma_U(m,x)
&=\E[Y\mid A=1,M=m,X=x,U=1]-\E[Y\mid A=1,M=m,X=x,U=0],\\
\delta_U(m,x)
&=\Pr(U=1\mid A=1,M=m,X=x)-\Pr(U=1\mid A=0,M=m,X=x).
\end{align*}
For a fixed mediator value, the observed outcome regression averages over the distribution of $U$ among treated individuals, while the cross-world mean $\theta_{10}$ uses the distribution of $U$ under control at that mediator value. Theorem 2 of \citet{vanderweele2010bias} gives
\[
\theta_{\mathrm{NIE}}-\theta_{\mathrm{obs}}
=\E_X\!\left[\int \gamma_U(m,X)\delta_U(m,X)\,dP(m\mid A=0,X)\right].
\]
Let $G$ and $D$ satisfy $|\gamma_U(m,x)|\leq G$ and $|\delta_U(m,x)|\leq D\leq 1$ for all $(m,x)$ in the relevant support. The absolute bias is then at most $GD$, because the integrand is bounded by the product of these two bounds and the mediator distribution integrates to one. Let $[\ell_\alpha,u_\alpha]$ denote the original interval in \eqref{eq:ci-calibrated}. For each hypothesized pair $(G,D)$ we report
\[
\bigl[\ell_\alpha-GD,\;u_\alpha+GD\bigr].
\]
This interval contains the NIE whenever the original interval contains $\theta_{\mathrm{obs}}$ and the bias bound holds. It first includes zero at $GD=\max(0,\ell_\alpha,-u_\alpha)$. All pairs use the same fitted model and original interval.

We generate 160 tables of size $n=500$, with 16 mechanisms per setting and two independent tables per mechanism. We draw $A$ and $U$ independently given $X$. The continuous mediator depends on $A$, $X$, and $U$, with additive Gaussian noise. The outcome depends on $A$, $X$, $M$, and $U$, with a treatment--mediator interaction and an additional quadratic mediator term in the nonlinear setting. We vary the coefficients of $U$ in the mediator and outcome equations to obtain five settings. The bias labels describe the direction and magnitude of $\theta_{\mathrm{obs}}-\theta_{\mathrm{NIE}}$. The MediPFN adapter, error head, and calibration are fixed. The generation mechanism determines the bounds used for each setting. Appendix~\ref{sec:sensitivity-simulation} gives the data-generating details and results.

\begin{figure}[!htbp]
\centering
\includegraphics[width=\linewidth]{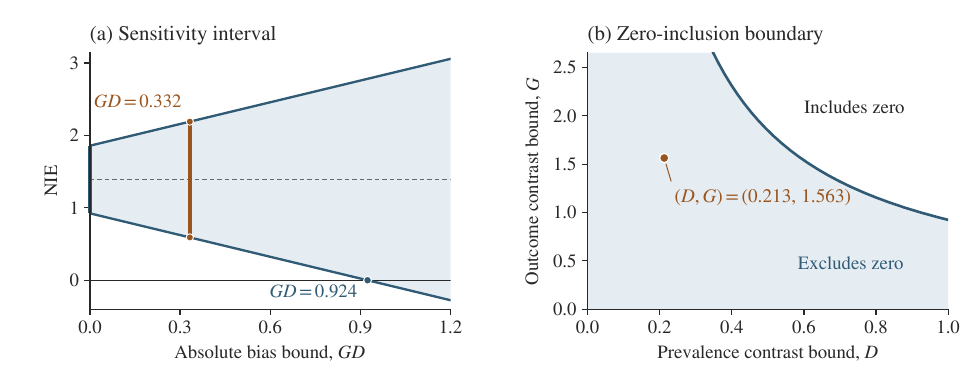}
\caption{Sensitivity to unmeasured mediator--outcome confounding. The dashed line marks the NIE estimate. Orange marks the selected bounds in both panels.}
\label{fig:sensitivity-my}
\end{figure}

Figure~\ref{fig:sensitivity-my} shows the adjustment for one table from the strong positive bias setting. The original interval is $[0.924,1.858]$. The bounds $G=1.563$ and $D=0.213$ are respectively the absolute outcome coefficient of $U$ and the maximum prevalence contrast under the generating mechanism (Appendix~\ref{sec:sensitivity-simulation}). The bias bound is 0.332 and the expanded interval is $[0.592,2.191]$. The interval first includes zero at $GD=0.924$. Panel (b) shows the pairs $(D,G)$ at which this happens. Extending the analysis to settings with unmeasured treatment--mediator or treatment--outcome confounding is a direction for future work.

\section{Discussion}
In this paper, we develop MediPFN for estimation and inference for natural indirect effects. The framework combines an adapted tabular foundation model with a cross-fitted influence-function-based estimator and learns standardized estimation-error quantiles across tasks. SFT, a symmetric interval, and independent calibration provide the uncertainty construction. The evaluated semi-synthetic settings show useful coverage on the studied covariate sources, while source-shift, small-sample, and heavy-tail findings illustrate the importance of assessing both coverage and interval length across mechanisms. Future work can expand the simulation priors, investigate estimation uncertainty under broader distribution shifts, and extend the framework to other causal inference settings such as instrumental-variable models.

\section*{AI Use Statement}
OpenAI Codex assisted with writing, literature searches, coding, and figure preparation. Responsibility for the final text, analyses, scientific conclusions, and references rests with the authors.

\begingroup
\fontsize{10}{11.5}\selectfont
\setlength{\bibsep}{0pt}
\bibliographystyle{apalike}
\bibliography{references,sensitivity_references}
\endgroup

\clearpage
\appendix
\linespread{1.07}\selectfont
\setlength{\parskip}{2pt plus 1pt}
\begin{center}
{\Large\bfseries\layoutrevision{Supplementary Material for}}\\[0.4em]
{\large\bfseries\layoutrevision{MediPFN: Amortized In-Context Nuisance Learning and Inference for Causal Mediation Analysis}}
\end{center}

\section{Implementation Details}
\label{sec:network-details}\label{sec:synthetic-banks}\label{sec:correction-details}\label{sec:runtime-details}
\noindent\currentchange{\textbf{Synthetic designs.} We sample from ten presets in eight DGP families to train the adapter, choosing the family uniformly and then a preset within the family. We first draw mechanism parameters, then generate a table conditional on those parameters. In the designs of \dgpref{cheng2026principal}, \dgpref{diaz2021nonparametric}, \dgpref{miles2020path}, and \dgpref{xia2023identification}, we set the coefficients on selected treatment-induced confounding paths to zero to satisfy the identification conditions in Section~\ref{sec:identification}. The full collection includes 14 presets from ten families.}

\noindent\textbf{DGP allocation and adaptations.} The training families draw on the designs of \citet{cheng2026principal}, \citet{diaz2020stochastic}, \citet{diaz2021nonparametric}, \citet{he2024adaptive}, \citet{jones2025mediation}, \citet{miles2020path}, \citet{shan2026mediation}, and \citet{zuo2025mediation}. The family based on \citet{jones2025mediation} contributes three training presets; its NNN preset is held out for composition-shift evaluation. The source-shift evaluation holds out the families based on \citet{tchetgen2012semiparametric} and \citet{xia2023identification} from mediation-specific fitting. Within a sampled family, a retained preset and its structural parameters are drawn before independently sampling the observations in a table. The simulated target is the superpopulation NIE in Section~\ref{sec:identification}, evaluated from the generating mechanism.

In our adaptation of \citet{cheng2026principal}, coefficients of the treatment-induced variable in the mediator and outcome equations are set to zero. For the intermediate-confounding design of \citet{diaz2021nonparametric}, we remove the induced variable and its interactions in the mediator and outcome equations. In our adaptation of \citet{miles2020path}, the induced covariates and their interactions in those equations are removed while the treatment--mediator interaction is retained. For the designs of \citet{xia2023identification}, we remove the induced-confounder effect in the outcome equation while retaining the mediator marginal and treatment--mediator interaction. Our adaptations of \citet{shan2026mediation} and \citet{zuo2025mediation} use complete-data variables before observation masks are applied.

\noindent\currentchange{\textbf{Adapter training and cross-fitting.} The backbone is fixed throughout training. The adapter cycles through the two propensity tasks and the outcome task with equal weight. It uses binary cross-entropy for the two propensity scores and squared error for the outcome, scaled by the outcome variance in the context rows with a floor of $10^{-6}$. Each of the 2,000 updates uses four independent tables of 128 rows. Selection on 16 development mechanisms gives the checkpoint at step 500.}

\currentchange{For each new table we use two treatment-stratified outer folds. An inner split of the context rows supplies out-of-fold outcome predictions as labels for the control-side nested regression. Each outer held-out fold supplies the score contributions for the NIE estimate. We clip the fitted propensity scores before forming the density ratio.} \thisrevision{For the learned intervals, we floor $\widehat s$ at $10^{-3}$ before standardizing errors and rescaling the interval radius.}

\begin{figure}[!tbp]
\centering
\includegraphics[width=0.9\linewidth]{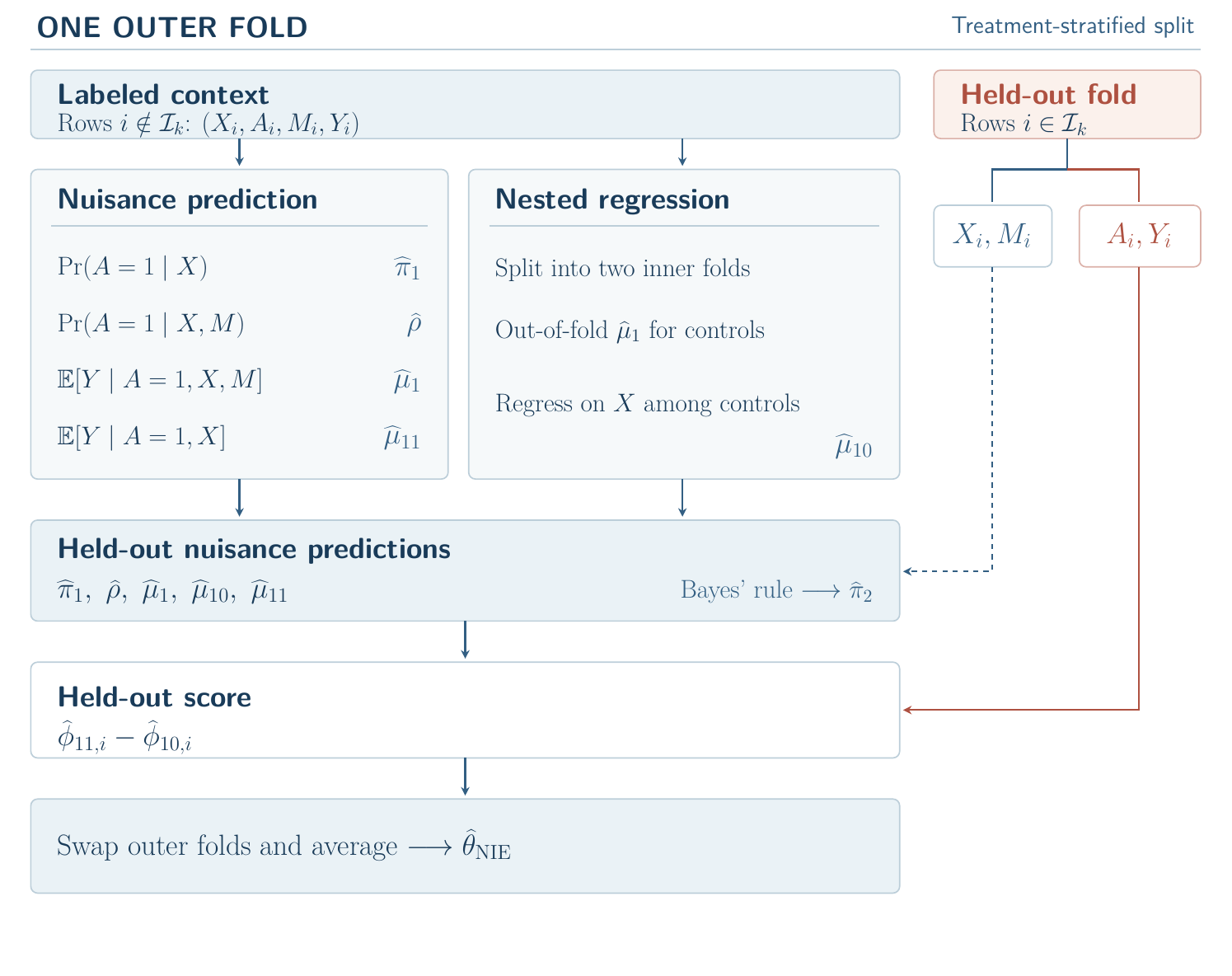}
\caption{\layoutrevision{Nested cross-fitting for one outer fold. The complementary rows supply the nuisance fits. The held-out observations enter the influence-function score through their nuisance predictions and observed values.}}
\label{fig:supp-crossfit}
\end{figure}

\layoutrevision{Figure~\ref{fig:supp-crossfit} shows how the inner split supplies labels for the nested regression and how the outer held-out rows enter the score.}

\noindent\currentchange{\textbf{Error quantile head.} The head receives a 1,024-dimensional pooled TabPFN representation and 14 summaries of the observed table and fitted estimator. We standardize the inputs with the synthetic training means and standard deviations, flooring the latter at $10^{-4}$. A layer normalization followed by two 256-unit GELU layers maps these features to six outputs $h_j(T)$. For $\mathcal T=\{0.005,0.025,0.05,0.95,0.975,0.995\}$ in increasing order, we construct
\[
 \widehat z_{\tau_j}(T)
 = h_1(T)+\sum_{k=2}^j\{\operatorname{softplus}(h_k(T))+10^{-4}\}.
\]
The positive increments keep the quantiles ordered. With $Z_i$ as in Section~\ref{sec:confidence-scaling}, the loss for table $i$ is
\[
 \ell_i=\frac{1}{6}\sum_{\tau\in\mathcal T}
       \ell_\tau\{Z_i-\widehat z_\tau(T_i)\}.
\]
We fit the head on 10,000 synthetic tables with mechanism-balanced weights and select its checkpoint on 1,024 development tables. We use AdamW with learning rate $10^{-3}$, batch size 128, and 6,000 updates.}

\noindent\textbf{Error-head inputs.} Let $p$ and $q$ denote the numbers of covariates and mediators, let $\hat\varphi_i$ be the cross-fitted NIE score for row $i$, and let $\widehat\sigma_Y$ denote the empirical standard deviation of $Y$ with denominator $n$. Put $s_Y=\max\{\widehat\sigma_Y,10^{-3}\}$. In the recorded feature order, the 14 summaries are
\[
\begin{aligned}
(&\log n,\ \log(1+p),\ \log q,\ \hat\theta/s_Y,\ \hat s/s_Y,\ \bar A,\ \bar Y/s_Y,\ \log s_Y,\\
&Q_{0.05}(\hat\varphi)/s_Y,\ Q_{0.50}(\hat\varphi)/s_Y,\ Q_{0.95}(\hat\varphi)/s_Y,\ c_{\pi},\ c_{\rho},\ \log(1+r_{\max})).
\end{aligned}
\]
Here $Q_t(\hat\varphi)$ is the empirical $t$-quantile of the scores, $c_{\pi}$ and $c_{\rho}$ are the averages across outer folds of the proportions of fitted treatment probabilities clipped in that fold, and $r_{\max}$ is the maximum fitted density ratio across folds. The two fitted probabilities are clipped to $[0.01,0.99]$ before the scores are computed. For each outer fold, the outcome-prediction query embeddings are summarized by their means and standard deviations across query rows and ensemble members; averaging the two fold summaries yields the 1,024-dimensional representation. The 14 values above are concatenated with this representation before standardization.

\noindent\currentchange{\textbf{Supervised fine-tuning.} We initialize SFT from this checkpoint and keep the adapter, backbone, and feature standardization fixed. Let $\mathcal I_{\mathrm{syn}}$ index the original synthetic tables and $\mathcal I_{\mathrm{real}}$ index 1,536 new tables generated using covariates from eight OpenML sources. With normalized weights $w_i$ for the synthetic set and $v_i$ for the real-covariate set, we minimize
\[
 \mathcal L_{\mathrm{SFT}}
 =\frac12\sum_{i\in\mathcal I_{\mathrm{syn}}}w_i\ell_i
  +\frac12\sum_{i\in\mathcal I_{\mathrm{real}}}v_i\ell_i.
\]
The weights $v_i$ give each OpenML source the same total weight. Each update samples 128 tables from each collection. We use 1,000 AdamW updates at learning rate $10^{-5}$ and select the checkpoint by the development pinball loss. To select the interval construction, we require at least 95\% coverage on each of two development sources (128 tables in total) and minimize source-balanced mean interval length among the eligible candidates.}

\noindent\currentchange{\textbf{Calibration.} The selected head supplies the two quantiles in the symmetric radius $r_\alpha(T)$. On 256 calibration tables from two separate OpenML sources, we compute the excess scores in \eqref{eq:cal-score}. The empirical ranks for 90\%, 95\%, and 99\% are 232, 245, and 255, respectively. We take the cumulative maximum of the three ordered corrections as coverage increases and add the resulting $c_\alpha$ to $r_\alpha(T)$ in \eqref{eq:ci-calibrated}.}

\noindent\currentchange{\textbf{Baseline implementation and timing.} The structural estimator uses a sandwich delta-method variance. RFF and RF use the empirical influence-function standard error to form Wald intervals. The two bootstrap baselines use 50 resamples for the coverage comparisons. Copies of the same subject share their fold assignments in the TabPFN V3 bootstrap. For timing, we compare MediPFN and the RF $g$-computation bootstrap with 100 resamples on 27 tables (12 real-covariate, nine synthetic, and six stress). Both procedures use the same CPU node and eight allocated threads. We time preprocessing, nuisance fitting, effect estimation, and interval construction after warm-up and summarize the paired time ratios by their median.}

\noindent\textbf{Nuisance learner settings.} The RF nuisance learners use 200 trees, a minimum leaf size of five, bootstrap sampling, and $\texttt{max\_features}=1.0$. The RFF learners standardize features using the support rows and concatenate the standardized features with 256 random Fourier features from an RBF kernel with coefficient $\gamma=1/d$, where $d$ is the input dimension. Their classification model uses logistic regression with $C=1$ and at most 1,000 iterations, and their regression model uses ridge regularization with coefficient one.

\section{Real Data}\label{sec:supp-data}
\currentchange{In this section, we describe the covariate populations and how we generate evaluation tables from them. The first set uses OpenML IDs 1468 (cnae-9), 1590 (adult), 188 (eucalyptus), and 40975 (car). The second uses IDs 1068 (pc1), 1120 (MagicTelescope), 1467 (climate-model-simulation-crashes), and 1480 (ilpd). Table~\ref{tab:real-sources} gives the dataset versions and dimensions after feature processing.}
\begin{table}[!htbp]
\centering\small
\begin{tabular}{clccc}
\toprule
Set & OpenML source & Version & Population rows & Covariates\\
\midrule
1 & cnae-9 (1468) & 1 & 1,080 & 30\\
1 & adult (1590) & 2 & 48,842 & 30\\
1 & eucalyptus (188) & 1 & 736 & 30\\
1 & car (40975) & 3 & 1,728 & 21\\
2 & pc1 (1068) & 1 & 1,109 & 21\\
2 & MagicTelescope (1120) & 1 & 19,020 & 10\\
2 & climate-model-simulation-crashes (1467) & 1 & 540 & 20\\
2 & ilpd (1480) & 1 & 583 & 11\\
\bottomrule
\end{tabular}
\caption{Observed covariate sources \reviewchange{used in the semi-synthetic evaluations.}}
\label{tab:real-sources}
\end{table}

\currentchange{We standardize columns within each covariate population, clip extreme standardized values, and sample rows with replacement. A sampled structural mechanism generates the treatment, mediator, and outcome. We average its indirect effect over the finite covariate population. The reference benchmark uses the same real-covariate design with 624 tables from 13 OpenML CC18 sources.}

\section{Additional Results}
In this section, we describe the evaluation summaries and present additional comparisons.

\medskip
\noindent\phantomsection\label{sec:additional-evaluation-details}\layoutrevision{\textbf{Evaluation details and paired summaries.}}\enspace
\thisrevision{The model and calibration are fixed before evaluation on the held-out tables. We report the held-out and reference results separately at $95\%$ nominal coverage unless otherwise stated. We weight the synthetic summaries equally across sample sizes, mechanism families, and presets within each family, and weight the real-covariate tables equally. Coverage is the weighted proportion of all tables whose intervals contain the true effect. For mean interval length, we renormalize the weights over reported intervals. All methods in the main synthetic and first OpenML comparisons returned intervals on every attempted table. Standard errors for coverage and paired differences cluster tables by mechanism. Ties share a rank, and interval lengths are marked only when at least two methods reach nominal coverage.}

For Table~\ref{tab:point-variance}, bias is the weighted mean estimation error. The reported SD is the square root of the weighted mean sampling variance within mechanism and sample size, with each variance estimated from two independent tables. The interval endpoints are weighted averages of the lower and upper endpoints over tables. Table~\ref{tab:robust-4m} summarizes MediPFN's evaluation. Coverage under composition shift is 91.67\%.
\begin{table}[!htbp]
\centering
\begin{tabular}{lccc}
\toprule
Setting & Tables & Coverage (\%) & Interval length\\
\midrule
ID & 768 & 94.01 & 0.607\\
Source shift & 288 & 92.97 & 1.752\\
Composition shift & 96 & 91.67 & 1.074\\
Real covariates, evaluation set 1 & 512 & \currentchange{94.14} & 0.554\\
Real covariates, evaluation set 2 & 512 & 95.90 & 3.656\\
\bottomrule
\end{tabular}
\caption{MediPFN interval performance at 95\% nominal coverage.}
\label{tab:robust-4m}
\end{table}

\medskip
\noindent\phantomsection\label{sec:additional-levels}\layoutrevision{\textbf{Additional nominal levels.}}\enspace
The error head returns quantiles for three fixed nominal levels. Table~\ref{tab:native-levels} reports the synthetic evaluation at 90\%, 95\%, and 99\%. A separate correction is calculated for each level from the independent calibration tasks. \currentchange{Tables~\ref{tab:real-sources-90} and \ref{tab:real-sources-99} give the per-source comparisons at 90\% and 99\%, respectively. Table~\ref{tab:stress-families-levels} reports MediPFN results by DGP design.}
\begin{table}[!htbp]
\centering\small
\begin{tabular}{llccc}
\toprule
Setting & Method & 90\% & 95\% & 99\%\\
\midrule
ID & MediPFN & 90.12 / 0.551 & 94.01 / 0.607 & 98.70 / 0.784\\
Source shift & MediPFN & 89.84 / 1.609 & 92.97 / 1.752 & 97.40 / 2.208\\
Composition shift & MediPFN & 89.58 / 0.999 & 91.67 / 1.074 & 94.79 / 1.352\\
\bottomrule
\end{tabular}
\caption{Coverage (\%) and interval length at three nominal levels on the synthetic evaluation.}
\label{tab:native-levels}
\end{table}

\thisrevision{Table~\ref{tab:real-calibration} compares intervals before and after calibration.}
\begin{table}[!htbp]
\centering\small
\begin{tabular}{ccc}
\toprule
Nominal level & Before calibration & After calibration\\
\midrule
90\% & 90.23 / 0.490 & 90.23 / 0.490\\
95\% & \thisrevision{93.16} / 0.530 & \currentchange{94.14} / 0.554\\
99\% & 96.29 / 0.603 & 97.46 / 0.668\\
\bottomrule
\end{tabular}
\caption{Coverage (\%) and interval length before and after calibration on the first semi-synthetic evaluation set.}
\label{tab:real-calibration}
\end{table}

\begin{table}[!htbp]
\centering\footnotesize
\setlength{\tabcolsep}{3pt}
\begin{tabular}{lcccccccc}
\toprule
Method & \multicolumn{2}{c}{cnae-9} & \multicolumn{2}{c}{adult} & \multicolumn{2}{c}{eucalyptus} & \multicolumn{2}{c}{car}\\
\cmidrule(lr){2-3} \cmidrule(lr){4-5} \cmidrule(lr){6-7} \cmidrule(lr){8-9}
 & $\widehat p_\alpha$ & $\bar\ell_\alpha$ & $\widehat p_\alpha$ & $\bar\ell_\alpha$ & $\widehat p_\alpha$ & $\bar\ell_\alpha$ & $\widehat p_\alpha$ & $\bar\ell_\alpha$\\
\midrule
MediPFN & 96.09 & \textbf{\underline{0.417}} & \textbf{\underline{89.06}} & 0.487 & \textbf{\underline{90.62}} & \textbf{\underline{0.652}} & \textbf{\underline{92.97}} & \textbf{\underline{0.498}}\\
Linear SEM & 84.38 & 0.251 & \underline{83.59} & 0.259 & \underline{88.28} & 0.241 & \underline{85.94} & 0.262\\
RFF  & \textbf{\underline{87.50}} & 1.835 & 81.25 & 2.801 & 73.44 & 2.195 & 79.69 & 1.002\\
RF  & \underline{86.72} & 0.920 & 76.56 & 0.510 & 82.03 & 0.665 & 79.69 & 0.516\\
RF bootstrap & 78.91 & 0.305 & 64.06 & 0.354 & 79.69 & 0.384 & 71.09 & 0.380\\
TabPFN V3 & 100.00 & \underline{2.536} & 99.22 & 5.551 & 100.00 & \underline{4.712} & 100.00 & \underline{6.382}\\
\bottomrule
\end{tabular}
\caption{Interval coverage and length by real-covariate source at 90\% nominal coverage.}
\label{tab:real-sources-90}
\end{table}

\begin{table}[!htbp]
\centering\footnotesize
\setlength{\tabcolsep}{3pt}
\begin{tabular}{lcccccccc}
\toprule
Method & \multicolumn{2}{c}{cnae-9} & \multicolumn{2}{c}{adult} & \multicolumn{2}{c}{eucalyptus} & \multicolumn{2}{c}{car}\\
\cmidrule(lr){2-3} \cmidrule(lr){4-5} \cmidrule(lr){6-7} \cmidrule(lr){8-9}
 & $\widehat p_\alpha$ & $\bar\ell_\alpha$ & $\widehat p_\alpha$ & $\bar\ell_\alpha$ & $\widehat p_\alpha$ & $\bar\ell_\alpha$ & $\widehat p_\alpha$ & $\bar\ell_\alpha$\\
\midrule
MediPFN & \textbf{\underline{99.22}} & \textbf{\underline{0.580}} & \underline{100.00} & \textbf{\underline{0.659}} & 95.31 & 0.866 & \underline{97.66} & 0.686\\
Linear SEM & 96.88 & 0.393 & 96.88 & 0.406 & 96.09 & 0.377 & \underline{97.66} & 0.411\\
RFF  & \underline{100.00} & \underline{2.874} & \textbf{\underline{98.44}} & 4.387 & 96.09 & 3.438 & 96.09 & 1.569\\
RF  & 96.09 & 1.441 & 94.53 & 0.798 & \underline{96.88} & 1.042 & 94.53 & 0.808\\
RF bootstrap & 84.38 & 0.429 & 77.34 & 0.485 & 88.28 & 0.529 & 85.16 & 0.540\\
TabPFN V3 & \underline{100.00} & 4.930 & \underline{100.00} & \underline{18.067} & \textbf{\underline{100.00}} & 9.020 & \textbf{\underline{100.00}} & 11.795\\
\bottomrule
\end{tabular}
\caption{Interval coverage and length by real-covariate source at 99\% nominal coverage.}
\label{tab:real-sources-99}
\end{table}

\begin{table}[!htbp]
\centering\small
\setlength{\tabcolsep}{4pt}
\begin{tabular}{lcccccc}
\toprule
\reviewchange{DGP source} & \multicolumn{2}{c}{90\%} & \multicolumn{2}{c}{95\%} & \multicolumn{2}{c}{99\%}\\
\cmidrule(lr){2-3} \cmidrule(lr){4-5} \cmidrule(lr){6-7}
& $\widehat p_\alpha$ & $\bar\ell_\alpha$ & $\widehat p_\alpha$ & $\bar\ell_\alpha$ & $\widehat p_\alpha$ & $\bar\ell_\alpha$\\
\midrule
\dgpref{cheng2026principal} & \currentchange{\textbf{\underline{89.58}}} & 1.894 & 93.75 & 2.126 & \textbf{\underline{100.00}} & 2.829\\
\dgpref{diaz2020stochastic} & \currentchange{\textbf{\underline{89.58}}} & 0.482 & 91.67 & 0.512 & 95.83 & 0.650\\
\dgpref{diaz2021nonparametric} & 91.67 & 0.997 & 93.75 & 1.048 & 97.92 & 1.279\\
\dgpref{tchetgen2012semiparametric} & 70.83 & 0.771 & 77.08 & 0.821 & 91.67 & 1.064\\
\dgpref{he2024adaptive} & \currentchange{\textbf{\underline{89.58}}} & 0.241 & 91.67 & 0.261 & 97.92 & 0.333\\
\dgpref{jones2025mediation}\reviewchange{, variant I} & 91.67 & 2.191 & \textbf{\underline{95.83}} & 2.281 & 97.92 & 2.761\\
\dgpref{jones2025mediation}\reviewchange{, variant II} & \currentchange{\textbf{\underline{89.58}}} & 1.804 & 93.75 & 1.892 & 97.92 & 2.324\\
\dgpref{miles2020path} & 95.83 & 0.863 & \textbf{\underline{95.83}} & 0.912 & \textbf{\underline{100.00}} & 1.170\\
\dgpref{shan2026mediation} & 93.75 & 2.062 & \textbf{\underline{95.83}} & 2.228 & \textbf{\underline{100.00}} & 2.834\\
\dgpref{xia2023identification}\reviewchange{, variant I} & 93.75 & 2.172 & 93.75 & 2.262 & 93.75 & 2.673\\
\dgpref{xia2023identification}\reviewchange{, variant II} & 91.67 & 9.872 & 93.75 & 10.699 & 97.92 & 13.582\\
\dgpref{zuo2025mediation} & 85.42 & 1.844 & 91.67 & 1.920 & 97.92 & 2.392\\
\bottomrule
\end{tabular}
\caption{MediPFN coverage (\%) and interval length by DGP design. Each design comprises 48 tables from 24 mechanisms, pooled over sample sizes and noise types. \currentchange{Bold underlining marks coverage within one percentage point of the nominal level.}}
\label{tab:stress-families-levels}
\end{table}

\begin{figure}[!tbp]
\centering
\includegraphics[width=\linewidth]{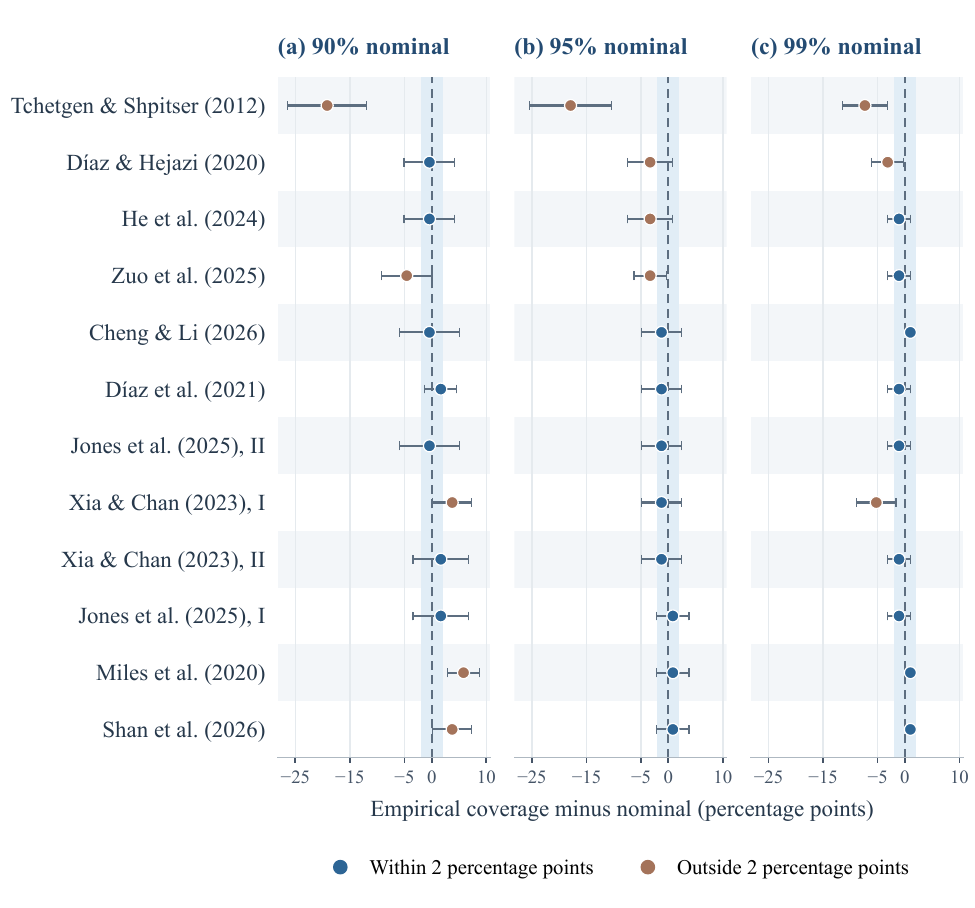}
\caption{\layoutrevision{Coverage minus nominal across the 12 stress-test designs. Blue marks coverage within two percentage points of nominal. Bars show one mechanism-clustered standard error on each side. Dashed lines mark zero.}}
\label{fig:supp-stress-dgp}
\end{figure}

\layoutrevision{Figure~\ref{fig:supp-stress-dgp} displays the coverage deviations by design and nominal level.}

\medskip
\noindent\phantomsection\label{sec:cpu-stress-results}\layoutrevision{\textbf{Heavy-tail and sample-size stress tests.}}\enspace
Table~\ref{tab:fresh-stress-subgroups} gives interval lengths and mechanism-clustered standard errors for selected stress subgroups.
\begin{table}[!htbp]
\centering
\begin{tabular}{lccc}
\toprule
Setting & Tables & Coverage (\%) & Interval length \\
\midrule
Additional $t_5$ noise & 192 & $93.75\, (1.80)$ & 3.046 \\
\reviewchange{Source shift} & -- & 92.36 & -- \\
\dgpref{tchetgen2012semiparametric} & 48 & $77.08\, (7.51)$ & 0.821 \\
\dgpref{he2024adaptive}\reviewchange{, added $t_5$ noise} & 16 & $87.50\, (8.84)$ & 0.207 \\
\bottomrule
\end{tabular}
\caption{Stress subgroups at 95\% nominal coverage. Parentheses give mechanism-clustered standard errors in percentage points.}
\label{tab:fresh-stress-subgroups}
\end{table}

\paragraph{Component-wise ablation.} On the 576 stress tables used for the pooled results in Figure \ref{fig:posttraining-ablation}, the original head's symmetric interval has 92.71\% coverage and mean length 2.298. The symmetric SFT interval before calibration has 94.10\% coverage and mean length 2.249. For both heads, symmetrization increases coverage by about nine percentage points and lengthens the intervals. The 95\% calibration expansion is zero, so the two symmetric SFT intervals coincide.

On the first semi-synthetic evaluation set, symmetrizing the original calibrated head's interval raises coverage from 76.37\% to 94.92\%, with mean length increasing from 0.402 to 0.639. For the fine-tuned head before calibration, the direct and symmetric intervals cover 86.33\% and \thisrevision{93.13\%}, with mean lengths of 0.370 and 0.530.

\medskip
\begin{samepage}
\noindent\phantomsection\label{sec:real-strata}\layoutrevision{\textbf{Real-covariate strata.}}\enspace
\currentchange{Table~\ref{tab:real-strata} summarizes a separate evaluation of the first real-covariate set, grouped by effect sign and sample size. Each evaluation table has a separately generated mechanism.}
\par\end{samepage}
\begin{table}[H]
\centering\small
\begin{tabular}{lccc}
\toprule
Stratum & Tables & Coverage (\%) & Length\\
\midrule
Positive effect & 208 & 93.27 & 0.609\\
Zero effect & 130 & 93.85 & 0.355\\
Negative effect & 174 & 92.53 & 0.566\\
$n=200$ & 166 & 88.55 & 0.769\\
$n=600$ & 178 & 94.38 & 0.486\\
$n=1000$ & 168 & 96.43 & 0.340\\
\bottomrule
\end{tabular}
\caption{MediPFN at 95\% nominal coverage on the first real-covariate evaluation set.}
\label{tab:real-strata}
\end{table}

\medskip
\noindent\layoutrevision{\textbf{Paired comparisons and additional levels.}}\enspace
\reviewchange{The 2,336-table reference benchmark} also permits comparison with the displayed externally fitted nuisance and structural methods in Table~\ref{tab:robust-rf}. These comparisons preserve the original table identities and folds; mean length is calculated only when an interval is produced. The SFT symmetric interval is evaluated using fixed estimator features; its computation time excludes nuisance fitting and point estimation. On the semi-synthetic component, MediPFN attains 95.03\% coverage and an interval length of 1.697.
\begin{table}[!htbp]
\centering\footnotesize
\setlength{\tabcolsep}{3pt}
\begin{tabular}{lcccccccc}
\toprule
Method & \multicolumn{2}{c}{ID} & \multicolumn{2}{c}{Source shift} & \multicolumn{2}{c}{Composition shift} & \multicolumn{2}{c}{Real covariates}\\
\cmidrule(lr){2-3} \cmidrule(lr){4-5} \cmidrule(lr){6-7} \cmidrule(lr){8-9}
 & $\widehat p_\alpha$ & $\bar\ell_\alpha$ & $\widehat p_\alpha$ & $\bar\ell_\alpha$ & $\widehat p_\alpha$ & $\bar\ell_\alpha$ & $\widehat p_\alpha$ & $\bar\ell_\alpha$\\
\midrule
MediPFN & \textbf{\underline{91.26}} & 0.591 & \textbf{\underline{92.97}} & 1.859 & \textbf{\underline{\currentchange{92.56}}} & 0.968 & \textbf{\underline{95.03}} & \textbf{\underline{1.697}}\\
Linear SEM & 79.53 & 0.359 & 62.50 & 1.212 & 71.87 & 0.615 & \underline{93.11} & 0.451\\
RFF  & 89.94 & 0.870 & \underline{90.10} & 1.669 & 85.42 & 1.413 & 91.99 & 6.032\\
RF  & 87.74 & 2.976 & 86.72 & 5.266 & 73.96 & 1.435 & 82.05 & 0.831\\
RF bootstrap & 59.79 & 1.566 & 47.14 & 1.788 & 14.58 & 0.671 & 78.04 & 0.545\\
TabPFN V3 & \underline{99.24} & 5.397 & 100.00 & 10.855 & \underline{100.00} & 6.662 & 100.00 & \underline{9.838}\\
\bottomrule
\end{tabular}
\caption{\reviewchange{Interval coverage and length on the reference benchmark.} Bold underlining marks first place and underlining marks second. Coverage is ranked by closeness to 95\% and length among methods reaching 95\%.}
\label{tab:robust-rf}
\end{table}

\begin{samepage}
For a paired coverage comparison, let $d_i$ be the difference between two coverage indicators on table $i$ and let $w_i$ be the prespecified normalized table weight. If $\mathcal I_g$ contains tables generated from mechanism $g$, the reported cluster standard error is
\begin{equation}
 \widehat\sigma_\Delta=\left[\frac{G}{G-1}\sum_{g=1}^G
 \left\{\sum_{i\in\mathcal I_g}w_i(d_i-\Delta\widehat p_\alpha)\right\}^2\right]^{1/2},
 \qquad \Delta\widehat p_\alpha=\sum_iw_i d_i.
\end{equation}
The same table, mechanism, and fold identities define each pair. A length difference uses jointly successful pairs and renormalizes their weights; its denominator can therefore differ from the coverage denominator. These standard errors describe variation across the sampled mechanisms in the specified evaluation and are not uncertainty estimates over arbitrary unobserved source populations.

The new-evaluation external comparisons, timing study, and newly generated stress evaluations use the table identities and folds specified above. As in the main text, source shift refers to the adaptation distribution; the foundation model's original training distribution is outside this experimental control.
\par\end{samepage}

\paragraph{Sensitivity analysis.}\label{sec:sensitivity-simulation}
We generate two independent tables of $n=500$ observations from each of 16 mechanisms in each of five settings. For every mechanism, the parameters below are sampled independently from the stated uniform ranges. We draw $X\sim\mathrm{Bernoulli}(p_X)$, $U\mid X\sim\mathrm{Bernoulli}\{\operatorname{expit}(u_0+u_X X)\}$, and $A\mid X\sim\mathrm{Bernoulli}\{\operatorname{expit}(a_0+a_X X)\}$, independently of $U$ conditional on $X$. The remaining variables follow
\begin{align*}
M&=\alpha_0+\alpha_X X+A(\alpha_A+\alpha_{AX}X)+\alpha_U U+\sigma_M\varepsilon_M,\\
Y&=\beta_0+\beta_X X+\beta_A A+(\beta_M+\beta_{AM}A)M+\beta_{M^2}M^2+\gamma U+\sigma_Y\varepsilon_Y,
\end{align*}
where $\varepsilon_M$ and $\varepsilon_Y$ are independent standard normal variables. The common sampling ranges are grouped as follows:
\begin{itemize}\setlength{\itemsep}{0pt}\setlength{\parskip}{0pt}
\item Covariate, confounder, and treatment: $p_X\in[0.35,0.65]$, $u_0\in[-0.55,0.10]$, $u_X\in[0.5,1.15]$, $a_0\in[-0.3,0.25]$, and $a_X\in[-0.5,0.6]$.
\item Mediator: $\alpha_0\in[-0.4,0.4]$, $\alpha_X\in[-0.45,0.45]$, $\alpha_A\in[0.55,1.05]$, $\alpha_{AX}\in[0.05,0.35]$, and $\sigma_M\in[0.8,1.2]$.
\item Outcome: $\beta_0\in[-0.4,0.4]$, $\beta_X\in[-0.5,0.65]$, $\beta_A\in[0.25,0.75]$, $\beta_M\in[0.45,0.9]$, $\beta_{AM}\in[0.15,0.4]$, and $\sigma_Y\in[0.8,1.25]$.
\end{itemize}
The setting-specific ranges for $(\alpha_U,\gamma,\beta_{M^2})$ are:
\begin{itemize}\setlength{\itemsep}{0pt}\setlength{\parskip}{0pt}
\item No confounding: $(0,[0.45,0.85],0)$.
\item Moderate positive bias: $([0.45,0.8],[0.45,0.85],0)$.
\item Strong positive bias: $([0.9,1.5],[1.05,1.6],0)$.
\item Strong negative bias: $([0.9,1.5],[-1.6,-1.05],0)$.
\item Nonlinear positive bias: $([0.75,1.2],[0.8,1.1],[0.12,0.28])$.
\end{itemize}
The no-confounding setting has $\alpha_U=0$ even though $U$ affects $Y$.

The data-generating mechanism gives the NIE from the conditional first and second moments of $M(1)$ and $M(0)$. For the bias-bound experiment, its known parameters give $G=|\gamma|$ and
\[
D=\max_{x\in\{0,1\}}\tanh\!\left(\frac{|\alpha_U(\alpha_A+\alpha_{AX}x)|}{4\sigma_M^2}\right).
\]
These are mechanism-specific oracle bounds. The estimator receives only $(X,A,M,Y)$, and the fitted model and calibration are held fixed across the evaluation.

Table~\ref{tab:sensitivity-my} reports coverage of the NIE. The original intervals cover the observed-data contrast in 155 of 160 tables (96.88\%), while they cover the NIE in 128 of 160. Using the mechanism-specific bounds, the expanded intervals cover the NIE in 158 of 160 tables. The mean length increases from 0.606 to 1.030. The largest changes occur in the two strong-confounding settings. In the no-confounding setting, $D=0$ and the two intervals coincide.

\begin{table}[H]
\centering
\small
\begin{tabular}{lrrrr}
\toprule
 & \multicolumn{2}{c}{Coverage (\%)} & \multicolumn{2}{c}{Interval length}\\
\cmidrule(lr){2-3}\cmidrule(lr){4-5}
Setting & Original & Expanded & Original & Expanded\\
\midrule
No confounding           & 96.88 &  96.88 & 0.629 & 0.629\\
Moderate positive bias   & 93.75 & 100.00 & 0.597 & 0.809\\
Strong positive bias     & 59.38 & 100.00 & 0.668 & 1.368\\
Strong negative bias     & 62.50 & 100.00 & 0.475 & 1.211\\
Nonlinear positive bias  & 87.50 &  96.88 & 0.660 & 1.135\\
\midrule
Pooled                   & 80.00 &  98.75 & 0.606 & 1.030\\
\bottomrule
\end{tabular}
\caption{NIE coverage and mean interval length at the 95\% nominal level. Each setting uses 32 tables. Expanded intervals use the bounds from the data-generating mechanism.}
\label{tab:sensitivity-my}
\end{table}

\section{Proof} \label{sec:coverage-proofs}
 In this section, we prove the coverage results in Section~3. All probabilities and expectations in the coverage proofs are conditional on the model fitted before the calibration stage.

\subsection{Proof of Proposition~\ref{prop:coverage}}
\begin{proof}
On a successful task, $\hat s>0$ and the endpoint definitions in \eqref{eq:ci} imply
\begin{equation}\label{eq:proof-cover-event}
\{\theta_{\mathrm{NIE}}\in C_{\alpha,0}(D_n)\}
=\{\hat z_\ell(T)\leq Z\leq\hat z_u(T)\}.
\end{equation}
Conditional on $T$ and $B=1$, continuity of $F$ and the ordering of the predicted quantiles therefore give
\begin{align}
&\left|\mathbb P\{\theta_{\mathrm{NIE}}\in C_{\alpha,0}(D_n)\mid T,B=1\}-(1-\alpha)\right|\nonumber\\
&\quad=\left|F(\hat z_u\mid T)-F(\hat z_\ell\mid T)-(1-\alpha)\right|
\leq\zeta(T).\label{eq:proof-conditional-cover}
\end{align}
\currentchange{Averaging \eqref{eq:proof-conditional-cover} over tasks with $B=1$ and multiplying by $\mathbb P(B=1)$ gives}
\begin{align}
&\left|\mathbb P\{\theta_{\mathrm{NIE}}\in C_{\alpha,0}(D_n)\}-(1-\alpha)\right|\nonumber\\
&\quad\leq\E[\zeta(T)\ind(B=1)]+(1-\alpha)\mathbb P(B=0).
\label{eq:proof-cover-bound}
\end{align}
Substituting $\mathbb P(B=0)=\delta$ proves \eqref{eq:coverage-bound}. Since $r_\alpha(T)$ is the larger absolute endpoint distance, the symmetric interval contains $C_{\alpha,0}(D_n)$ on every successful task. Its coverage is therefore at least the lower bound obtained from \eqref{eq:proof-cover-bound}.
\end{proof}

\subsection{Proof of Corollary~\ref{cor:proper-rate}}
\begin{proof}
For $\tau\in\{\alpha/2,1-\alpha/2\}$, let $q_\tau(T)$ be the conditional $\tau$-quantile of $Z$ given $T,B=1$, and let $\hat q_\tau(T)$ be its prediction. Write $\ell_\tau(u)=u\{\tau-\ind(u<0)\}$ and define the conditional pinball risk $L_\tau(q;T)=\E[\ell_\tau(Z-q)\mid T,B=1]$. By continuity of $F$,
\begin{equation}\label{eq:proof-pinball}
L_\tau(\hat q_\tau;T)-L_\tau(q_\tau;T)
=\int_{q_\tau}^{\hat q_\tau}\{F(t\mid T)-\tau\}\,dt.
\end{equation}
The integral is nonnegative in either orientation. The lower and upper density bounds between $q_\tau$ and $\hat q_\tau$ imply, respectively,
  \begin{align}
L_\tau(\hat q_\tau;T)-L_\tau(q_\tau;T)
&\geq \frac{c}{2}(\hat q_\tau-q_\tau)^2,\label{eq:proof-risk-lower}\\
|F(\hat q_\tau\mid T)-\tau|
&\leq C|\hat q_\tau-q_\tau|.\label{eq:proof-cdf-upper}
\end{align}
Let $\mathcal R_\tau$ denote the expectation of the excess risk in \eqref{eq:proof-risk-lower}, conditional on $B=1$. By the definition of $\mathcal R$, $\mathcal R=(\mathcal R_{\alpha/2}+\mathcal R_{1-\alpha/2})/2$. Applying Cauchy--Schwarz to \eqref{eq:proof-risk-lower}--\eqref{eq:proof-cdf-upper} gives
  \begin{align}
\E[\zeta(T)\mid B=1]
&\leq C\sum_{\tau}\sqrt{\E[(\hat q_\tau-q_\tau)^2\mid B=1]}\nonumber\\
&\leq C\sum_{\tau}\sqrt{2\mathcal R_\tau/c}
\leq 2C\sqrt{2\mathcal R/c}.\label{eq:proof-quantile-error}
\end{align}
Applying Proposition~\ref{prop:coverage} and $\mathbb P(B=1)=1-\delta$, we obtain
  \begin{align}
\mathbb P\{\theta_{\mathrm{NIE}}\in[\hat\theta_{\mathrm{NIE}}\pm\hat s r_\alpha(T)]\}
&\geq (1-\delta)\{1-\alpha-\E[\zeta(T)\mid B=1]\}\nonumber\\
&\geq (1-\alpha)(1-\delta)-2C\sqrt{2\mathcal R/c}.
\end{align}
This proves \eqref{eq:risk-bound}; the rate follows by substituting the stated bound for $\mathcal R$.
\end{proof}

\subsection{Proof of Theorem~\ref{thm:calibrated-coverage}}
\begin{proof}
Let $c_\alpha^0$ be the $k$th ordered calibration score. Assumption~  \ref{ass:calibration-sampling}  implies  $c_\alpha^0<\infty$ almost surely. \currentchange{As $S_*=+\infty$ if $B=0$, $S_*\leq c_\alpha^0$ implies $B=1$} and, since $c_\alpha\geq c_\alpha^0$, $\theta_{\mathrm{NIE},*}\in C_\alpha(D_*;\mathcal C_m)$. It remains to lower-bound $\mathbb P(S_*\leq c_\alpha^0)$.

For $s\in\mathbb R\cup\{+\infty\}$, define
\[
N_s=\sum_{i=1}^m\ind(S_i<s),\qquad
\overline K_m(s-)=\frac{1}{m}\sum_{i=1}^m\mathbb P(S_i<s),\qquad
K(s-)=\mathbb P(S_*<s).
\]
The strict inequalities account for ties: $c_\alpha^0<s$ if and only if $N_s\geq k$. By the definition of $\Delta_m$, taking left limits gives $\overline K_m(s-)\leq K(s-)+\Delta_m$. The same inequality holds at $s=+\infty$ by a limit. Since the calibration scores are independent, Hoeffding's inequality for the indicators in $N_s$ gives
\begin{equation}\label{eq:proof-heterogeneous-tail}
\mathbb P(c_\alpha^0<s)
\leq\exp\!\left[-2m\left\{\frac{k}{m}-\Delta_m-K(s-)\right\}_+^2\right].
\end{equation}

We next integrate \eqref{eq:proof-heterogeneous-tail} over the independent target score. Let $U=K(S_*-)$. For $u\in(0,1)$, let $s_u$ be the generalized $u$-quantile of $S_*$ on the extended real line. Then $K(s_u-)\leq u$ and $\mathbb P(S_*\leq s_u)\geq u$. Consequently,
\begin{equation}\label{eq:proof-rank-domination}
\mathbb P(U\leq u)\geq \mathbb P(S_*\leq s_u)\geq u.
\end{equation}
Thus $U$ is stochastically dominated by a uniform random variable, including when $S_*$ has atoms or a mass at $+\infty$.

Set $a=k/m-\Delta_m$. When $a>0$, the right side of \eqref{eq:proof-heterogeneous-tail} is nondecreasing in $K(s-)$. Using the independence of $S_*$ and the calibration scores, followed by \eqref{eq:proof-rank-domination}, yields
\begin{align}
\mathbb P(c_\alpha^0<S_*)
&\leq\E\!\left[\exp(-2m(a-U)_+^2)\right]\nonumber\\
&\leq 1-a+\int_0^a e^{-2mt^2}\,dt
\leq 1-a+\sqrt{\frac{\pi}{8m}}.\label{eq:proof-gaussian-integral}
\end{align}
For $a\leq0$, the zero lower bound is immediate. Combining \eqref{eq:proof-gaussian-integral} with $\mathbb P\{\theta_{\mathrm{NIE},*}\in C_\alpha(D_*;\mathcal C_m)\}\geq \mathbb P(S_*\leq c_\alpha^0)$ proves \eqref{eq:calibration-bound}.
\end{proof}
\end{document}